\documentclass[journal, twocolumn, 10pt]{IEEEtran}
\usepackage{bm}
\usepackage{caption}
\usepackage{multirow}
\usepackage{eucal}
\usepackage{bbm}
\usepackage{diagbox} 
\usepackage{cite}
\usepackage{booktabs}
\usepackage{graphicx}
\usepackage{subfig}

\ifCLASSINFOpdf
\else
\fi
\usepackage{amsmath}

\usepackage{amsfonts} 
\usepackage{amssymb} 
\usepackage{amsthm}
\theoremstyle{definition}

\newtheorem{remark}{Remark}

\usepackage{mathtools}
\allowdisplaybreaks

\usepackage{algorithmic}

\usepackage{algorithm}

\usepackage{subfig}

\begin{document}

% Industrial Anomaly Localization Jointly Using Weakly Supervised Learning and Rule Knowledge
%Integrating knowledge-driven and data-driven approaches to modeling
\title{Reconstruction-Based Anomaly Localization via Knowledge-Informed Self-Training
	\thanks{Manuscript received ---. This work was supported by the Zhejiang Provincial Natural Science Foundation of China (Grant No. LY23F020018), the Key-Area Research and Development Program of Guangdong Province (Grant No. 2021B0101410004),  and the National Program for Special Support of Eminent Professionals.}}
\author{Cheng~Qian, Xiaoxian~Lao, Chunguang~Li,~\IEEEmembership{Senior~Member,~IEEE}
	\thanks{The authors are with the College of Information Science and Electronic Engineering, Zhejiang University, Hangzhou 310027, China. C. Qian and C. Li are also with the Ningbo Research Institute, Zhejiang University, Ningbo 315100, China (C. Li is the corresponding author, email: cgli@zju.edu.cn). }}
	
%\markboth{IEEE Transactions on Neural Networks and Learning Systems}%
\markboth{}%
{Qian \MakeLowercase{\textit{et al.}}: Reconstruction-Based Anomaly Localization via Knowledge-Informed Self-Training}

\maketitle

\begin{abstract}
Anomaly localization, which involves localizing anomalous regions within images, is a significant industrial task. Reconstruction-based methods are widely adopted for anomaly localization because of their low complexity and high interpretability. Most existing reconstruction-based methods only use normal samples to construct model. If anomalous samples are appropriately utilized in the process of anomaly localization, the localization performance can be improved. However, usually only weakly labeled anomalous samples are available, which limits the improvement. In many cases, we can obtain some knowledge of anomalies summarized by domain experts. Taking advantage of such knowledge can help us better utilize the anomalous samples and thus further improve the localization performance. In this paper, we propose a novel reconstruction-based method named knowledge-informed self-training (KIST) which integrates knowledge into reconstruction model through self-training. Specifically, KIST utilizes weakly labeled anomalous samples in addition to the normal ones and exploits knowledge to yield pixel-level pseudo-labels of the anomalous samples. Based on the pseudo labels, a novel loss which promotes the reconstruction of normal pixels while suppressing the reconstruction of anomalous pixels is used. We conduct experiments on different datasets and demonstrate the advantages of KIST over the existing reconstruction-based methods.

\end{abstract}

\begin{IEEEkeywords}
	Anomaly localization, image reconstruction, knowledge, self-training.
	%Distributed estimation, distributed algorithm, diffusion, quantization, adaptive combiner.
\end{IEEEkeywords}

% \fi
\IEEEpeerreviewmaketitle

\section{Introduction}
%\IEEEPARstart{W}{ith} the rapid development of smart manufacturing, lots of intelligent methods have been applied to various industrial processes, such as quality inspection \cite{QualityInspection1, QualityInspection2, QualityInspection3, QualityInspection4}, fault diagnosis \cite{FaultDiagnosis1, FaultDiagnosis2, FaultDiagnosis3}. Deep learning-based anomaly localization is essential for industrial surface quality inspection.

\IEEEPARstart{A}{nomaly} localization, generally referring to localizing anomalous regions within images, is a significant task emerging in different industrial scenarios~\cite{QualityInspection1, QualityInspection2, QualityInspection3, QualityInspection4, FaultDiagnosis1, FaultDiagnosis2, FaultDiagnosis3}. In many cases of anomaly localization, normal samples are convenient to acquire while the anomalous ones with fine-grain pixel-level annotations are hard to obtain, which makes anomaly localization challenging.

In the literature, a major portion of anomaly localization methods, including reconstruction-based methods, generative model-based methods, etc~\cite{CAE0, CAE1, SkipGan, SkipConnect, TexNet, FeatClusterCAE, PyramialNet, PatchReconstruct, MemAE, FeatEditAdvNet, LatentResample, GausDescrip, AdpPryGrap, SSIMAE, WFDL, RIAD}, utilize only normal samples to construct model. Among them, the reconstruction-based method is most popular for its low computational cost and high interpretability. The insight of the reconstruction-based method is that the model trained by normal samples can only reconstruct normal pixels well. The pixels that can not be reconstructed well are likely to be anomalous. In the reconstruction-based method, generally, a neural network named encoder first compresses the input image to a low-dimensional feature. Then, the low-dimensional feature will be mapped by another network named decoder to get the reconstructed image. Further, the reconstruction residual image is computed and used to localize anomalies.

Many reconstruction-based anomaly localization methods have been proposed.
Typically, Baur et al.~\cite{CAE0} comes up with the vanilla convolutional autoencoder (CAE) for unsupervised anomaly localization with brain MR images. Youkachen et al.~\cite{CAE1} employ CAE for industrial image reconstruction to localize anomalous regions in  hot-rolled strips. Some researchers utilize other neural networks for industrial anomaly localization~\cite{SkipGan, SkipConnect, TexNet, FeatClusterCAE, PyramialNet}. The above methods reconstruct the whole image for the localization, and there also exists methods  that reconstruct the image patch-wise~\cite{PatchReconstruct}. 
To learn the parameters of the neural networks, various loss functions~\cite{SSIMAE, RIAD, WFDL} have been utilized. The existing reconstruction-based methods employing neural networks have made fruitful progress in anomaly localization. However, it is often observed that the neural networks also reconstruct the anomalous pixels well, which influences the localization performance. 
This is largely because that the reconstruction ability of the neural network is so powerful that it reproduces the input regardless of whether the input is normal or anomalous. To restrict the reconstruction ability of the neural networks, several methods have been proposed, such as memory banks~\cite{MemAE}, feature clustering~\cite{FeatClusterCAE, FeatEditAdvNet}, feature modeling~\cite{LatentResample, GausDescrip, AdpPryGrap}. 
%These methods leverage normal samples to constrain the distribution of the features of both normal and anomalous samples. Since only normal samples are used, the distribution of the anomalous features may not be constrained appropriately. 
% Besides, there are some works attempting to improve the loss function such as \cite{SSIMAE, RIAD, WFDL}. Since only normal samples are used, the localization performance is not improved significantly compared with the original loss.
The above methods only use normal samples to construct the reconstruction model. If anomalous samples are also used for the construction of the model, better localization performance can be expected. In many cases, only a few weakly labeled anomalous samples with image-wise annotations are available. The image-level labels limit the benefit of the anomalous samples.

It is noticed that even for some previously unseen anomalies, human experts are able to spot and localize them. This is largely because the experts have absorbed some experiential knowledge of anomalies. The knowledge could bring valuable information about the anomalies.
Taking advantage of such knowledge  can help make better use of the weakly labeled anomalous samples and thus further improve the localization performance. In many cases, we can actually acquire some knowledge of anomalies summarized by domain experts. How to utilize the knowledge is worth considering.  

Since the knowledge is usually expressed by natural language with vagueness, we represent it in the form of fuzzy rules to make it convenient to be utilized. Further, we propose a novel reconstruction-based method named knowledge-informed self-training (KIST). In the KIST, after constructing the initial model using only normal samples, two stages are iteratively conducted. In the first stage, pixel-wise pseudo-labels of anomalous samples are generated by a fuzzy rule-based pseudo-label producing module. In the second stage, based on the normal samples and the anomalous samples with the pseudo-labels, a contrastive-reconstruction loss is used to update the model. With the iterations in the manner of self-training, the quality of the pseudo-labels and the performance of the model are gradually improved. The main novelty and contribution are listed as follows.

\begin{itemize}
	\item [1)] We propose KIST framework that utilizes knowledge to make better use of the weakly labeled anomalous samples and thus improve the performance of anomaly localization. KIST can be applied to a large portion of reconstruction models. It has good effectiveness and generality.
	\item [2)] We use fuzzy rules to represent the knowledge of anomalies and then quantitatively evaluate the anomaly degree of regions of anomalous samples. Based on the evaluation, we produce pixel-wise pseudo-labels of anomalous samples to update the model.
	\item [3)]  We use a contrastive-reconstruction loss, which promotes the reconstruction of normal pixels while suppressing the reconstruction of anomalous pixels for model updating.
\end{itemize}

The remaining of this article is organized as follows. In Section II, we introduce how to evaluate the anomaly degree of regions of anomalous samples based on the knowledge, and propose the   KIST method. In Section III, we demonstrate the advantages of the KIST using qualitative and quantitative experimental results. We conclude the paper in Section IV.

\section{KIST Method}
	
	\subsection{Overview of KIST}
\begin{figure*}[htbp]
		\centering
		\includegraphics[width=7in]{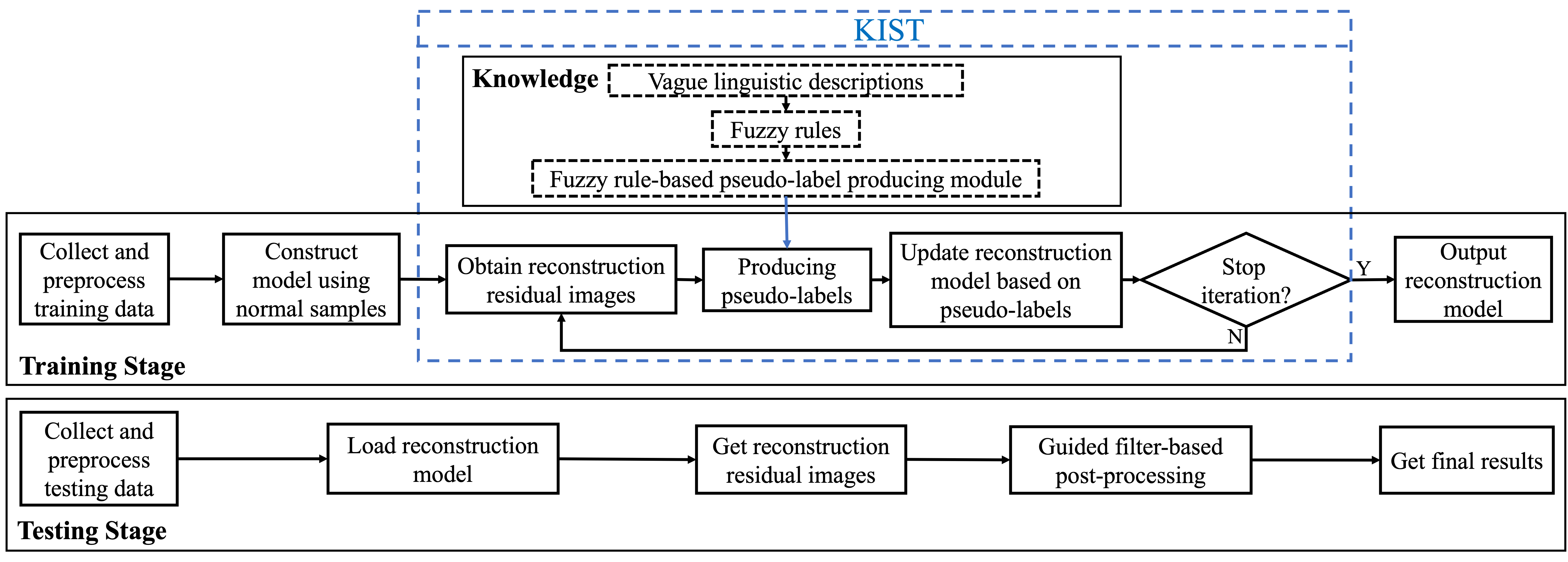}
		\caption{Flowchart of KIST-based anomaly localization scheme.}
		\label{Fig:flowchart_KIST}
	\end{figure*}

	Besides obtaining some anomalous samples with image-level annotations, we can actually acquire some knowledge of anomalies described by domain experts. However, how to take advantage of such knowledge is worth considering. 
	In this section, we propose a novel reconstruction-based method named knowledge-informed self-training (KIST), which integrates knowledge into reconstruction model in the manner of self-training. The flowchart of the KIST-based anomaly localization scheme is shown in Fig.~\ref{Fig:flowchart_KIST}. 
	We represent the knowledge by fuzzy rules.	In the training stage, firstly the initial model is constructed by only normal samples. Next, pixel-wise pseudo-labels of anomalous samples are generated by a fuzzy rule-based pseudo-label producing module.
	Based on the normal samples and the anomalous ones with the pseudo labels, a contrastive-reconstruction loss which promotes the reconstruction of normal pixels while suppressing the reconstruction of anomalous pixels is used to update the reconstruction model.
	The two stages, pseudo-label producing and model updating, are iteratively conducted in the manner of self-training so that the quality of the pseudo-labels and the performance of the model are gradually improved. 
%	In testing stage, we input the image and obatin the reconstruction residual image. Then the guided filter-based post-processing method is employed to smooth the high residual regions caused by backgroud noise. 
	
%	To this end, we propose to use knowledge to get more information about anomalous regions from weakly labeled anomalous samples. To be specific, we propose KIST framework, which can be illustrated in Fig.\ref{Fig:flowchart_KIST}. In the training stage. the model is pretrained only by normal samples. Then, rule knowledge is loaded and membership grade of anomaly is introduced to measure the anomalous degree of the region. After that, two iterative steps are performed. First, we use the latest model and knowledge to produce pseudo-labels from weakly labeled anomalous samples. Second, the latest pseudo-labels is employed to update model in a fully-supervised way. The two steps performed in an iterative manner to make the pseudo-labels and the model mutually optimized. When testing, the guided filter-based post-processing method is present to smooth the high residual regions caused by noise in the backgroud. Details of the KIST is introduced in the rest of this section.
	
	\begin{figure}[htbp]
		\centering
		\includegraphics[width=3.5in, height=3.2in]{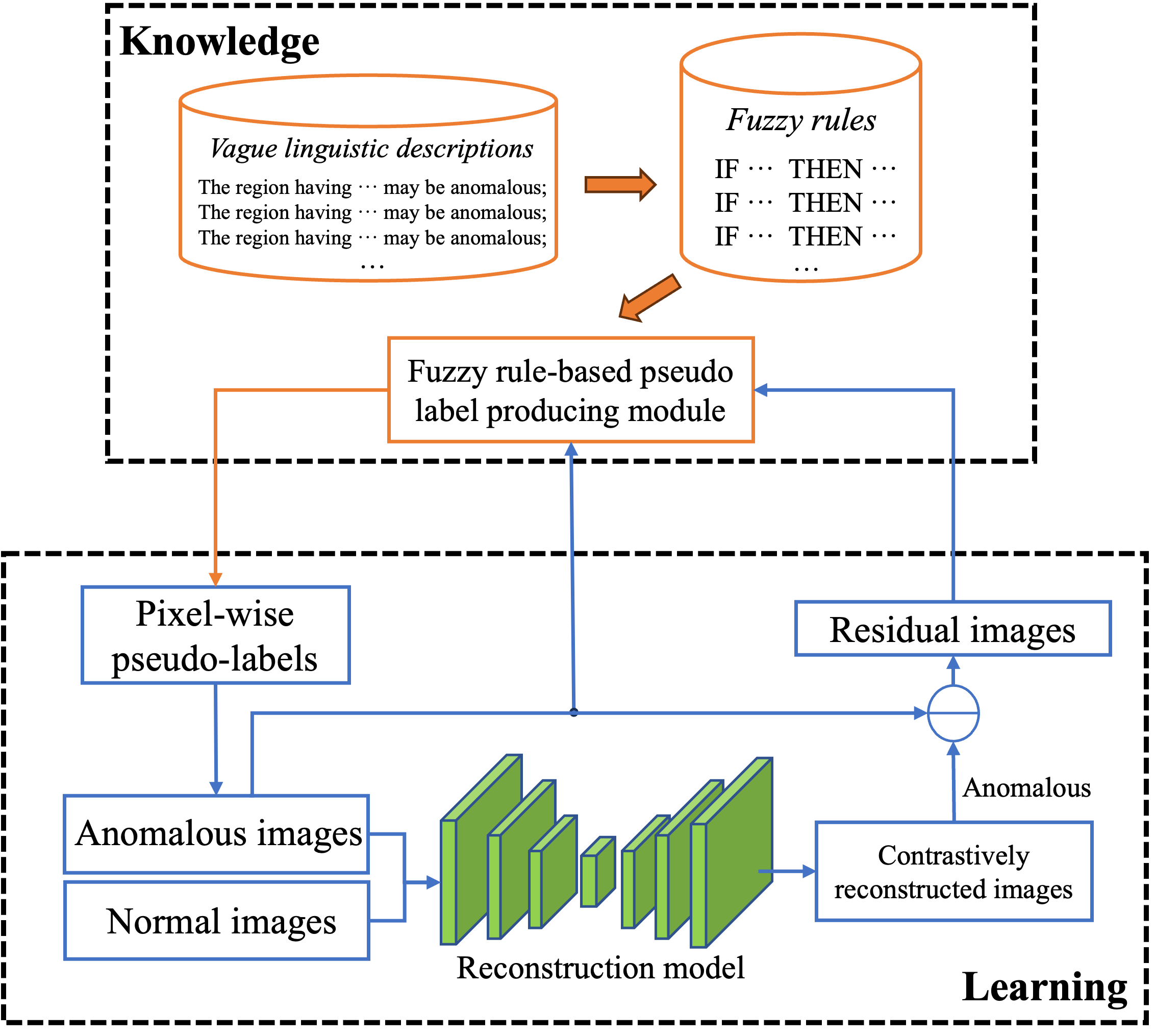}
		\caption{Diagram of KIST.}
		\label{Fig:KIST_Module}
	\end{figure}
	
	\subsection{Knowledge-Based Anomaly Grade Evaluation}
	\subsubsection{Knowledge representation using fuzzy rule}
\begin{figure}[htbp]    % 常规操作\begin{figure}开头说明插入图片
			% 后面跟着的[htbp]是图片在文档中放置的位置，也称为浮动体的位置，关于这个我们后面的文章会聊聊，现在不管，照写就是了
			\centering            % 前面说过，图片放置在中间
			\captionsetup[subfloat]{labelsep=none,format=plain,labelformat=empty}
			\subfloat[(a)]
			{
				\includegraphics[width=0.8in]{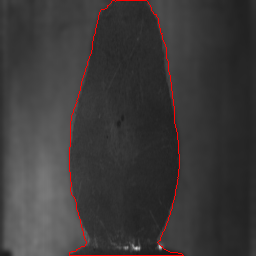}
				\hspace{-2mm}
			}
			\subfloat[(b)]
			{
				\includegraphics[width=0.8in]{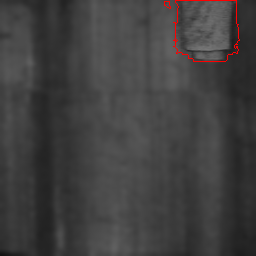}
				\hspace{-2mm}
			}
			\subfloat[(c)]
			{
				\includegraphics[width=0.8in]{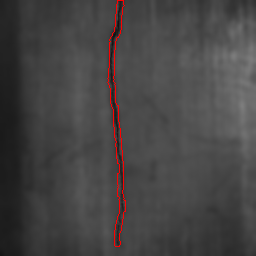}
				\hspace{-2.5mm}
			}
			\subfloat[(d)]
			{
				\includegraphics[width=0.8in]{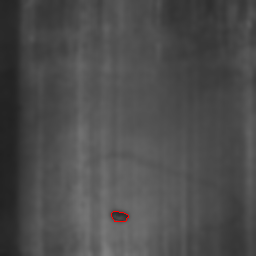}
			}
			\caption{Examples of anomalous samples (the anomalous regions are contoured with red line). The anomalous region of (a) has low gray value and large area. The anomalous region of (b) has high gray value and rectangle shape. The anomalous region of (c) has low gray value and slender shape. The anomalous region of (d) has low gray value and small area.}
			\label{fig:Knowledge_example}            
		\end{figure}

	In many cases, we may obtain some knowledge about anomalies summarized by domain experts. However, it is usually difficult for the experts to represent the knowledge in precise terms. Instead, vague linguistic descriptions are often used such as
	\begin{itemize}
		\item The region having low gray value is anomalous;
		\item The region having very large area or very small area may be anomalous;
		\item The region having rectangle shape may be anomalous;
		\item The region having slender shape may be anomalous;
	\end{itemize}
	The corresponding examples are shown in Fig.\ref{fig:Knowledge_example}.
	
		As we know, fuzzy rule is a useful tool to handle vague and non-numerical information. In this subsection we use fuzzy rules to represent the vague linguistic descriptions.
		
		We begin with the simplest case. Take the proposition ``The region having low gray value  may be anomalous" as an example. It can be represented by the fuzzy rule
		\begin{equation}\label{Eq:fr_1r1v_example}
			\begin{aligned}
				&\text{IF the gray value of region is low} \\
				&\text{THEN the region is anomalous}
			\end{aligned}
		\end{equation}
		 where the fuzzy propositions ``the gray value of region is low" and ``the region is anomalous" are named as antecedent and consequent, respectively. Strictly speaking, a fuzzy rule is associated with a truth value $tv$~\cite{FuzzyBook}, and normally the value of $tv$ is set to 1 by default. Here, considering that ``may be" is used by the experts in the linguistic description, we set $0<tv < 1$. 
		To express (\ref{Eq:fr_1r1v_example}) more generally, we denote by $\mathcal{V}(\mathbf{r})$ a property of the region $\mathbf{r}$, $F$ a fuzzy set corresponding to a fuzzy predicate such as ``low" and ``high", and $F_{an}$ also a fuzzy set.
		Thus fuzzy rules (\ref{Eq:fr_1r1v_example}) with truth value $tv$ can be expressed as 
		\begin{equation}\label{Eq:fr_1r1v}
			\begin{aligned}
				&\text{IF $\mathcal{V}(\mathbf{r})$ is $F$} \\
				&\text{THEN $\mathbf{r}$ is $F_{an}$ ($tv$)} 
			\end{aligned}
		\end{equation}
		Specifically, in (\ref{Eq:fr_1r1v_example}) $\mathcal{V}(\mathbf{r})$ is corresponding to the gray value of the region $\mathbf{r}$, and $F$ and $F_{an}$ are fuzzy sets corresponding to the predicates ``low" and ``anomalous", respectively.			
		
		In many cases, the acquired knowledge is more complicated, and more than one property is mentioned in a fuzzy rule. For example, the fuzzy rule corresponding to the description ``The region having low gray value and small area may be anomalous" involves two properties, namely gray value and area. Assuming a fuzzy rule with the truth value $tv$ has $n$ properties $\mathcal{V}_k$ along with the corresponding fuzzy sets $F_k$ for $1\le k \le n$. 
%		Since the consequent ``$\mathbf{r}$ is $F_{an}$" is true only if each elemental proposition ``$\mathcal{V}_k(\mathbf{r})$ is $F_k$" is true.
		We use AND operator to connect different elemental propositions. The fuzzy rule can be expressed as 
		\begin{equation}
			\begin{aligned}
				&\text{IF \!$\mathcal{V}_1(\mathbf{r}) $ is $F_1$ AND $\mathcal{V}_2(\mathbf{r})$ is $F_2$  AND\! $\cdots$ \!AND $\mathcal{V}_n(\mathbf{r})$ is $F_n$}\\
				&\text{THEN $\mathbf{r}$ is $F_{an}$ ($tv$)} \nonumber
			\end{aligned}
		\end{equation}
		Moreover, when there are $m$ fuzzy rules available, assuming the $i$th$(1\le i \le m)$ rule with the truth value $tv^{(i)}$ involves $n^{(i)}$ properties $\mathcal{V}_j^{(i)}$ along with the corresponding fuzzy sets $F_j^{(i)}$ for $1\le j \le n^{(i)}$, the $i$th fuzzy rule can be represented as
%		Since the consequent of each rule is the same, we use propositional connective OR to associate compound propositions of antecedent of each rule, then the fuzzy rule can be represented as
		\begin{equation}\label{Eq:fuzzy_mrules2}
		\begin{aligned}
			&\text{IF $\mathcal{V}_1^{(i)}(\mathbf{r})$ is $F_1^{(i)}$ AND $\mathcal{V}_2^{(i)}(\mathbf{r})$ is $F_2^{(i)}$ AND $\! \cdots \!$ AND}\\
			&\quad \text{ $\mathcal{V}_j^{(i)}(\mathbf{r})$ is $F_j^{(i)}$ AND $\! \cdots \!$ AND $\mathcal{V}_{n^{(i)}}^{(i)}(\mathbf{r})$ is $F_{n^{(i)}}^{(i)}$}\\
			&\text{THEN $\mathbf{r}$ is $F^{(i)}_{an}$ ($tv^{(i)}$)}
		\end{aligned}
	\end{equation}
%		\begin{equation}\label{Eq:fuzzy_mrules2}
%		\begin{aligned}
%			&\text{IF ($\mathcal{V}_1^{(1)} $ is $F_1^{(1)}$ AND $\mathcal{V}_2^{(1)}$ is $F_2^{(1)}$   $\cdots$ AND $\mathcal{V}_{n_1}^{(1)}$ is $F_{n_1}^{(1)}$)}\\
%			&\text{THEN $\mathbf{r}$ is $F_{an}$. ($tv^{(1)}$)}\\
%			&\text{IF ($\mathcal{V}_1^{(2)} $ is $F_1^{(2)}$ AND $\mathcal{V}_2^{(2)}$ is $F_2^{(2)}$   $\cdots$ AND $\mathcal{V}_{n_2}^{(2)}$ is $F_{n_2}^{(2)}$)}\\
%			&\text{THEN $\mathbf{r}$ is $F_{an}$. ($tv^{(2)}$)}\\
%			&\cdots\\
%			&\text{IF ($\mathcal{V}_1^{(m)} $ is $F_1^{(m)}$ AND $\mathcal{V}_2^{(m)}$ is $F_2^{(m)}$   $\cdots$  AND $\mathcal{V}_{n_m}^{(m)}$ is $F_{n_m}^{(m)}$)},\\
%			&\text{THEN $\mathbf{r}$ is $F_{an}$. ($tv^{(m)}$)}
%		\end{aligned}
%		\end{equation}

%		\begin{equation}\label{Eq:fuzzy_mrules2}
%			\begin{aligned}
%				&\text{IF ($\mathcal{V}_1 ^ {(1)} $ is $F_1^{(1)}$ AND $\mathcal{V}_1^{(2)}$ is $F_1^{(2)}$   $\cdots$  $\mathcal{V}_1^{(n_1)}$ is $F_1^{(n_1)}$)}\\
%				&\text{OR ($\mathcal{V}_2 ^ {(1)} $ is $F_2^{(1)}$ AND $\mathcal{V}_2^{(2)}$ is $F_2^{(2)}$   $\cdots$  $\mathcal{V}_2^{(n_2)}$ is $F_2^{(n_2)}$)}\\
%				&\cdots\\
%				&\text{OR ($\mathcal{V}_y ^ {(1)} $ is $F_m^{(1)}$ AND $\mathcal{V}_y^{(2)}$ is $F_m^{(2)}$   $\cdots$  $\mathcal{V}_y^{(n_m)}$ is $F_m^{(n_m)}$)},\\
%				&\text{THEN $\mathbf{r}$ is $F_{an}$}.
%			\end{aligned}
%		\end{equation}
			\subsubsection{Anomaly grade evaluation}
			After representing knowledge using the fuzzy rules as (\ref{Eq:fuzzy_mrules2}), we give quantitative evaluations of how anomalous a region is based on these rules. In (\ref{Eq:fuzzy_mrules2}), $F_{an}$ is a fuzzy set. It corresponds to a membership function $F_{an}(\cdot)$ which outputs the grade of membership of the input in the fuzzy set ``anomalous". For a specific region $\mathbf{r}$, its membership grade in ``anomalous" is $F_{an}(\mathbf{r})$.  The higher the membership grade of $\mathbf{r}$ in the fuzzy set ``anomalous", the more likely $\mathbf{r}$ being anomalous. Since the value of $F_{an}(\mathbf{r})$ can reflect the degree of anomaly of the region $\mathbf{r}$, we use $F_{an}(\mathbf{r})$ as an indicator of the anomaly degree, and define the anomaly grade of the region $\mathbf{r}$ as $g(\mathbf{r}) \triangleq F_{an}(\mathbf{r})$.
			 In the following, we describe how to calculate $F_{an}(\mathbf{r})$ and thus get  $g(\mathbf{r})$.

			%	As we can see in (\ref{Eq:fuzzy_mrules2}), $F_{an}(\mathbf{r})$ depends on the degree of truth $T(p_c)$ of consequent $p_c$, where we use $F_{an}(\mathbf{r})=T(p_c)$. Clearly, $T(p_c)$ equals the degree of truth $T(p_a)$ of antecedent $p_a$. Therefore, to obtain $F_{an}(\mathbf{r})$ of $\mathbf{r}$, $T(p_a)$ of (\ref{Eq:fuzzy_mrules2}) should be acquired.
			
			We begin with the simple case where there are only one rule given as (\ref{Eq:fr_1r1v}).
			Given a region $\mathbf{r}$, we denote by $v $ the value of $\mathcal{V}(\mathbf{r})$, where $v$ is taken from a universal set $V$.
			For example, the gray value $v_g = \mathcal{V}_g(\mathbf{r})$ of $\mathbf{r}$ is taken from the gray value set $V_g=\{0,1,\cdots, 255\}$.  To get $F_{an}(\mathbf{r})$, we need to evaluate the truth value of the antecedent  ``$v$ is $F$" of (\ref{Eq:fr_1r1v}). Let $p_a$ denote the antecedent and  $T(p_a)$  its  truth value. $T(p_a)$ can be obtained by first calculating  the membership grade $F(v)$ of $v$ in the fuzzy set $F$ and letting
			\begin{equation}
				T(p_a) = F(v). \nonumber
			\end{equation}
			After obtaining the truth value of the antecedent $T(p_a)$, we can acquire the truth value $T(p_c)$ of the consequent $p_c$ ``$\mathbf{r}$ is $F_{an}$" by jointly considering $T(p_a)$ and the truth value of the rule $tv$.
			Since $tv$ can be considered as the possibility of the rule (\ref{Eq:fr_1r1v}) being valid, we use probabilistic operator~\cite{FuzzyBook} to connect $tv$ and $T(p_a)$ and further get $T(p_c)$ as
			\begin{equation} 
				T(p_c)=tv * T(p_a). \nonumber
			\end{equation}
			Letting the membership grade $F_{an}(\mathbf{r})$ of $\mathbf{r}$ in $F_{an}$ equals to the truth value of the consequent ``$\mathbf{r}$ is $F_{an}$", we have 
			\begin{equation} 
				F_{an}(\mathbf{r})=T(p_c). \nonumber
			\end{equation} 
			Finally, the anomaly grade $g(\mathbf{r})$ is given by $g(\mathbf{r}) = F_{an}(\mathbf{r})$.
			
			Next, we consider the more general case where there are $m$ rules and each rule is given as (\ref{Eq:fuzzy_mrules2}). We first derive $F^{(i)}_{an}(\mathbf{r})$ of the $i$th rule.  
			To get $F^{(i)}_{an}(\mathbf{r})$, we need to evaluate the truth value of the antecedent of (\ref{Eq:fuzzy_mrules2}). We denote by $p_a^{(i)}$ the antecedent of (\ref{Eq:fuzzy_mrules2}), and $T(p_a^{(i)})$ its true value. In $ p_a^{(i)}$, the AND operater associates the elemental propositions ``$v_j^{(i)} $ is $F_j^{(i)}$"  where $v_j^{(i)}=\mathcal{V}_j^{(i)}(\mathbf{r})$ for $ j \in[1,n^{(i)}] $. Let $p_e^{(i,j)}$ denote the elemental proposition ``$v_j^{(i)} $ is $F_j^{(i)}$" and $T(p_e^{(i, j)})$ denote the truth value of $p_e^{(i,j)}$. Then $ p_a^{(i)}$ can be written as ``$p_e^{(i, 1)}$ AND $p_e^{(i, 2)}$ AND $\cdots$ AND $p_e^{(i, n_i)}$". To compute the truth value $T(p_a^{(i)})$ of $ p_a^{(i)}$, we use the formula defined by Zadeh~\cite{FuzzySet}
			\begin{equation} 
				T(p_e^{(i, 1)}\;\text{AND}\;p_e^{(i, 2)}\;\text{AND}\;\cdots\; \text{AND}\; p_e^{(i, n_i)})= \wedge_{j=1}^{n^{(i)}} T(p_e^{(i, j)}) \nonumber
			\end{equation} 
			 and get
			\begin{equation} 
				T(p_a^{(i)})= \wedge_{j=1}^{n^{(i)}} T(p_e^{(i, j)}), \nonumber
			\end{equation} 
			where $\wedge$ is the minimum operator. Since $p_e^{(i,j)}$ is the elemental proposition ``$v_j^{(i)} $ is $F_j^{(i)}$", its truth value $T(p_e^{(i, j)})$ can be acquired by calculating the membership grade of $v_j^{(i)}$ in the fuzzy set  $F_j^{(i)}$, i.e.,
			\begin{equation}
				T(p_e^{(i, j)}) = F_j^{(i)}(v_j^{(i)}). \nonumber
			\end{equation}
			Thus we have 
			\begin{equation} 
				T(p_a^{(i)}) =\wedge_{j=1}^{n^{(i)}} F_j^{(i)}(v_j^{(i)}).\nonumber
			\end{equation}
			After obtaining the truth value of the antecedent $T(p_a^{(i)})$, we can acquire the truth value $T(p_c^{(i)})$ of the consequent $p_c^{(i)}$ ``$\mathbf{r}$ is $F^{(i)}_{an}$" by jointly considering $T(p_a^{(i)})$ and the truth value $tv^{(i)}$ of the $i$th rule as
			\begin{equation}\label{eq:pc_irule}
				T(p_c^{(i)}) = tv^{(i)} * T(p_a^{(i)}).
			\end{equation}
			Letting the truth value of the consequent ``$\mathbf{r}$ is $F^{(i)}_{an}$" equal to the membership grade $F^{(i)}_{an}(\mathbf{r})$ of $\mathbf{r}$ in $F^{(i)}_{an}$, we have
			\begin{equation}
				F_{an}^{(i)}(\mathbf{r}) = T(p_c^{(i)}),
			\end{equation}
			and the corresponding anomaly grade $g^{(i)}(\mathbf{r})$ is given as 
			\begin{equation}
				g^{(i)}(\mathbf{r}) = F_{an}^{(i)}(\mathbf{r}).
			\end{equation}
			When there are $m$ fuzzy rules available for the evaluation of how anomalous the region $\mathbf{r}$ is, we can obtain $m$ anomaly grade $g^{(i)}(\mathbf{r})$ for $i \in [1, m]$. We obtain the anomaly grade $g(\mathbf{r})$ by jointly considering the different $g^{(i)}(\mathbf{r})$. Since $\mathbf{r}$ is likely to be anomalous as long as it gets a high anomaly grade according to one fuzzy rule,  we take the highest grade among $g^{(i)}(\mathbf{r})$ for $i \in [1, m]$ as the anomaly grade of $\mathbf{r}$, i.e.,
			\begin{equation}\label{eq:ms_an}
				g(\mathbf{r})=\vee_{i=1}^m g^{(i)}(\mathbf{r}).
			\end{equation}
			where $\vee$ is the maximum operator. 
	
	\subsection{Knowledge-Informed Self-Training}
	The knowledge can help us evaluate how anomalous a given region is. Taking advantage of such knowledge, we may produce fine-grained annotations for weakly labeled anomalous samples. How to utilize the information brought by the knowledge to improve the performance of anomaly localization is worth considering. As we know, the self-training methods enrich training data by assigning pseudo-labels to unlabeled or weakly labeled samples for better training models. In light of the idea of self-training,  we propose a knowledge-informed self-training (KIST) method for reconstruction-based anomaly localization. The diagram of KIST is shown in Fig.~\ref{Fig:KIST_Module}. As one can see, pixel-wise pseudo-labels of anomalous samples are produced based on fuzzy rule-based pseudo-label producing module. Based on normal samples and the anomalous ones with the pseudo-labels, a contrastive-reconstruction loss which promotes the reconstruction of normal pixels while suppressing the reconstruction of anomalous pixels is used to update the model. The pseudo-label producing stage and the model updating stage are iteratively conducted in the manner of self training so that the quality of the pseudo-labels and the performance of the model are gradually improved. 
	
	\subsubsection{Reconstruction model based on neural network}
%	 We expect to produce the pseudo-labels that highlight the anomalous pixels. As we know, reconstruction-based methods construct reconstruction models using only normal samples. The pixels that are reconstructed badly and have high reconstruction residuals are likely to be anomalous.

	In the reconstruction-based method, generally, an encoder $E$ first compresses the input image $\mathbf{x}$ to a low-dimensional feature $\mathbf{z}$. Then, $\mathbf{z}$ is mapped by a decoder $D$ to get the reconstructed image $\hat{\mathbf{x}}$.  The above reconstruction process is given as
	 \begin{gather}
	 	\mathbf{z} = E(\mathbf{x}),\nonumber\\
	 	\hat{\mathbf{x}} =D(\mathbf{z}). \nonumber
	 \end{gather}
	 After the reconstruction, the residual image can be computed and used to localize anomalies. It is expected that the reconstruction model can only reconstruct the normal pixels well such that the pixels having high residuals are considered anomalous.
	 
	 Generally, the encoder $E$ and decoder $D$ can be implemented using neural networks. There are various types of neural networks. Among them, the convolutional neural network, which employs convolutional layers to reduce the parameters and help learn the spatial information, could efficiently process image data. Thus, we adopt convolutional neural networks as basic structures of the reconstruction model.
	 
	 \subsubsection{Initial model constructing using normal samples}
	 We construct the initial reconstruction model $G$, which consists of the encoder $E$ and the decoder $D$, using only normal samples. The set of normal samples is denoted by $X_n = \left\{\mathbf{x}_1, \mathbf{x}_2, \cdots, \mathbf{x}_{N_n} \right\}$.
	  Let $\Theta$ denote the parameters of the model $G$. We expect the reconstructed image $\hat{\mathbf{x}}=G(\mathbf{x};\Theta)$ to be close to $\mathbf{x}$. To this end, $\Theta$ can be determined by minimizing
	 	\begin{equation}\label{eq:l2_loss}
	 		L(\Theta) =\frac{1}{N_n} \sum_{i=1}^{N_n} \Vert \hat{\mathbf{x}}_i  - \mathbf{x}_{i} \Vert_2^2.
	 \end{equation}
	\subsubsection{Pseudo-label producing for anomalous samples}
	After constructing the initial model, we could obtain a residual image $e(\mathbf{x})$ defined by
	\begin{gather} 
		e(\mathbf{x}) \triangleq \lvert \hat{\mathbf{x}}  - \mathbf{x} \rvert^2 . \nonumber
	\end{gather}
	If $\mathbf{x}$ is an anomalous sample with image-wise annotation,  relying on the knowledge-based anomaly grade evaluation mechanism and the residual image $e(\mathbf{x})$ of $\mathbf{x}$, we can generate a pixel-wise pseudo-label for $\mathbf{x}$. In the following we describe how to generate the pseudo-label for $\mathbf{x}$.
	
	If using a residual threshold to binarize the residual image of the image $\mathbf{x}$, we could obtain a set of regions whose residuals are greater than the threshold. Since anomalous regions could be included in any of those regions, we evaluate each of them to find the regions that are likely to be anomalous. We name the region that are likely to be anomalous as anomalous-like region. Once all anomalous-like regions in the image $\mathbf{x}$ are acquired, the pseudo-label of $\mathbf{x}$ can be acquired. To be specific, the pseudo-label producing procedures are as follows. 
	
%	First, we traverse the residual threshold and acquire all anomalous-like regions. Note that different anomalous-like regions might be found based on different thresholds. After all anomalous-like regions are acquired, they would be combined and used to produce the pseudo-label. 

	We traverse residual threshold $t$ in a residual threshold set
	\begin{equation}\label{eq:ts}
		T_e = \left\{t | t=\mu_e+ ns\sigma_e, \lceil 1/s \rceil \le n \le \lfloor 3/s \rfloor, n \in \mathbb{N} \right\},
	\end{equation}
	where $\mu_e$ and $\sigma_e$ are the mean and standard deviation of the residuals of all the pixels of the normal training samples, and $s\sigma_e$  is the step size. Pixels whose residual is larger than $t$ are suspected to be anomalous. The traversal of $t$ begins at $\mu_e+ \sigma_e$ to exclude most of the normal pixels that have relatively small residuals and stops at $\mu_e+ 3\sigma_e$ to include anomalous pixels as much as possible. Note that the total number of steps is determined by the parameter $s$. In practice we can adjust $s$ to make the total number of step smaller than $10$.
	For each residual threshold $t \in T_e$, we binarize $\mathbf{x}$ using $t$ and get a set of connected regions $R_{t,\mathbf{x}} = \{\mathbf{r}^{(1)}, \mathbf{r}^{(2)}, \cdots, \mathbf{r}^{(N_t)}\}$ with residuals larger than $t$. Then, to find anomalous-like regions, we evaluate the anomaly degree of each region $\mathbf{r}^{(i)} \in R_{t,\mathbf{x}}$ using the anomaly grade $g(\mathbf{r}^{(i)})$ introduced in the last subsection. If $g(\mathbf{r}^{(i)})$ is high enough,  $\mathbf{r}^{(i)}$ is likely to be an anomalous region. Therefore, we set an anomaly grade threshold $\alpha$ and consider the region whose anomaly grade is higher than $\alpha$ to be an anomalous-like region. By doing this, for each residual threshold $t$, we could acquire an anomalous-like region set 
	\begin{equation}\label{eq:find_an_regions}
		R_{t, \mathbf{x}}^* = \{ \mathbf{r} | g(\mathbf{r}) \ge \alpha, \forall \mathbf{r} \in R_{t, \mathbf{x}}\}.
	\end{equation}
	The anomalous-like region set $R_\mathbf{x}$ of the image $\mathbf{x}$ is given as the union of all $R_{t, \mathbf{x}}^*$, which is expressed as
	\begin{equation}\label{eq:update_an_set}
		R_{\mathbf{x}} = \bigcup_t R_{t, \mathbf{x}}^*, 
	\end{equation}
	where $R_\mathbf{x} = \{\mathbf{r}^{(1)}, \cdots, \mathbf{r}^{(N_\mathbf{x})}\}$ includes all anomalous-like regions in $\mathbf{x}$. 
	After $R_\mathbf{x}$ is acquired, all anomalous-like regions $\mathbf{r}^{(i)} \in R_{\mathbf{x}}$ would be used to produce the pseudo-label $\mathbf{PL}_\mathbf{x}$ of the image $\mathbf{x}$. The pseudo-label of $\mathbf{x}$ is given by
	\begin{gather}\label{eq:pseudo_label}
		\mathbf{PL}_\mathbf{x}(p) =
			\begin{cases}
				1 ,& \quad p \in \mathbf{r} \text{ for } \mathbf{r} \in R_\mathbf{x}\\
				0 ,& \quad \text{otherwise},
			\end{cases}
	\end{gather}

\begin{figure*}[htbp]
		\centering
		\includegraphics[width=2\columnwidth, height=5.4in]{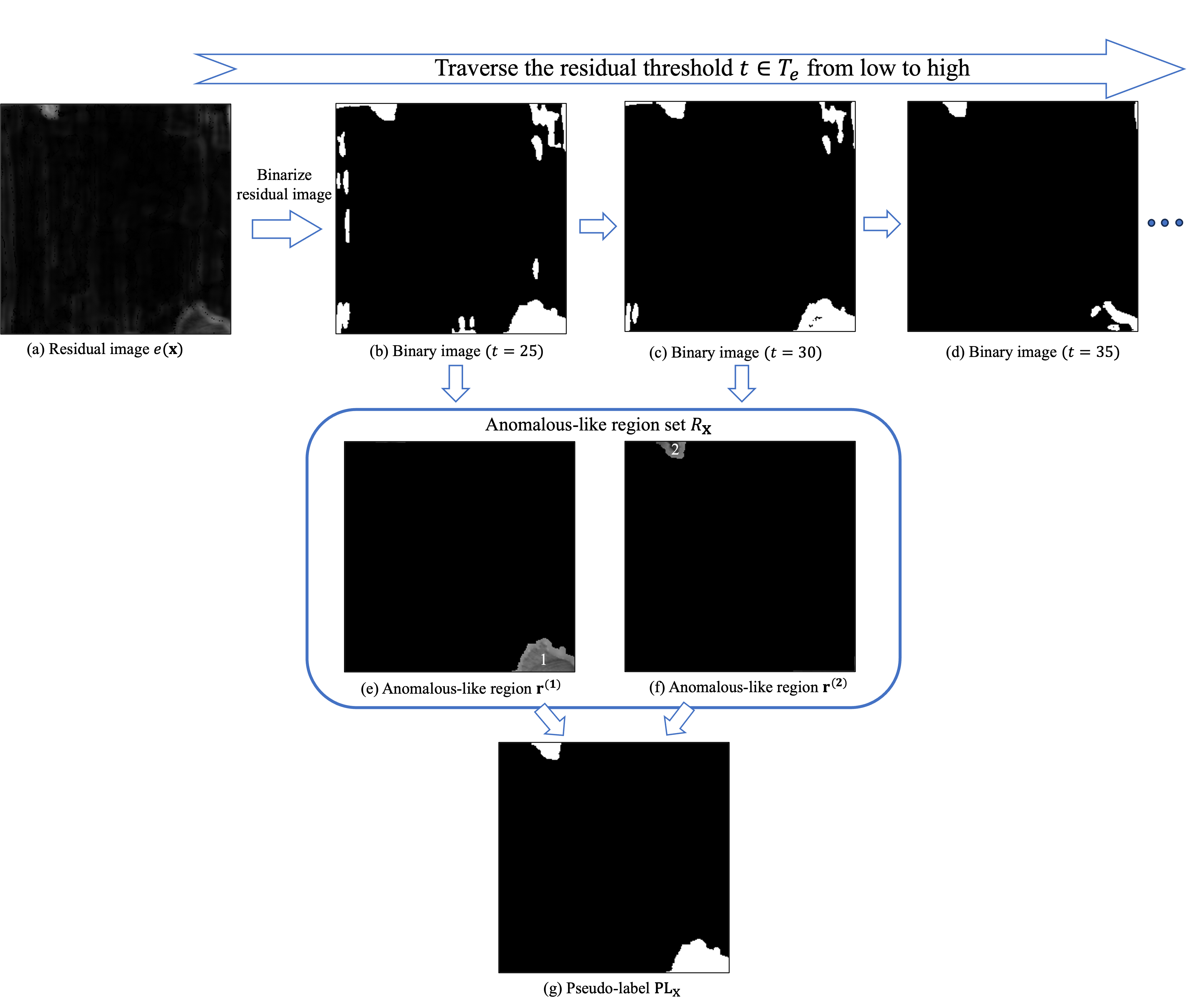}
		\caption{Diagram of pseudo-label producing process.  }
		\label{Fig:diagram_pl_producing}
	\end{figure*}

We present a diagram in  Fig.~\ref{Fig:diagram_pl_producing}  to demonstrate the overall pseudo-label producing process. After the pseudo-labels of all anomalous samples are produced, we use them to update the reconstruction model.

%	For each threshold $t\in T_e$, we binarize the residual image $e(\mathbf{x})$ and acquire a set of regions with residuals higher than $t$. Then, based on the fuzzy rules and the original image $\mathbf{x}$, we calculate the anomaly grade $g(\mathbf{r})$ of each region $\mathbf{r}$ and find the anomalous-like regions whose $g(\mathbf{r})$ is greater than $\alpha$. After obtaining all anomalous-like regions, we produce the pseudo-label $\mathbf{PL}_{\mathbf{x}}$.
	
%	In a word, when an anomalous image is given, the pseudo-label producing module would regard the whole image as a region at first. As the traversal of threshold, the image would be split into many regions where each of them includes at most one anomalous region. Then, the pseudo-label of each region is produced individually. After that, the pseudo-label of the whole image is obtained by combining the pseudo-labels of all sub-regions. Once the pseudo-labels of all anomalous images are produced, they will be used to refresh the training data.
	
	 \subsubsection{Model updating by contrastive reconstruction}
	 To achieve good anomaly localization performance, a reconstruction model should reconstruct normal pixels well and reconstruct anomalous pixels badly.  Here we jointly utilize the normal samples and the anomalous samples with pixel-level pseudo-labels to achieve this purpose. Specifically, a contrastive-reconstruction loss, which promotes the reconstruction of normal pixels while suppressing the reconstruction of anomalous pixels, is employed to update the model based on the pixel-wise labels. Given $N$ training samples $X$ including $N_n$ normal images  $X_{n}$ and $N_{an}$ anomalous images $X_{an}$  along with their pseudo-labels $\mathbf{PL}_{\mathbf{x}_i}$ for ${\mathbf{x}_i}\in X_{an}$, the parameter $\Theta$ of the model can be determined by minimizing the following contrastive-reconstruction loss
 	\begin{equation}\label{eq:contrastive_reconstruction_loss}
	 	\begin{aligned}
	 		L_{\text{cr}}(\Theta) & =\frac{1}{N_n*W*H} \sum_{\mathbf{x}_i\in X_{n}} \Vert \hat{\mathbf{x}}_i  - \mathbf{x}_{i} \Vert_2^2 \\
	 		& -\lambda\frac{1}{\sum_{\mathbf{x}_i\in X_{an}} \Vert \mathbf{PL}_{\mathbf{x}_i} \Vert_0} \sum_{\mathbf{x}_i\in X_{an}} \left(  \Vert \mathbf{PL}_{\mathbf{x}_i} \circ (\hat{\mathbf{x}}_i  - \mathbf{x}_{i})  \Vert_2^2 \right)
	 	\end{aligned}
	 \end{equation}
%	\begin{equation}\label{eq:contrastive_loss}
%			\begin{aligned}
%			L_{c r}(\Theta) & =\frac{1}{N} \sum_{i=1}^{N} \sum_{r=0}^{H-1} \sum_{c=0}^{W-1}\left[\left(1-\mathbf{PL}_{\mathbf{x}_{i}}(r, c)\right) \left(\mathbf{x}_{i}(r, c) - \hat{\mathbf{x}}_i(r,c)\right)^2\right. \\
%			& \left.-\lambda\left(\mathbf{PL}_{\mathbf{x}_{i}}(r, c)\right) \left(\mathbf{x}_{i}(r, c) - \hat{\mathbf{x}}_i(r,c) \right)^2\right],
%		\end{aligned}
%	\end{equation}
	where $H$ and $W$ are the height and width of $\mathbf{x}$, $\lambda$ is a hyperparameter, $\Vert \mathbf{PL}_{\mathbf{x}_i} \Vert_0$ is the number of non-zero entries in $\mathbf{PL}_{\mathbf{x}_i}$, and $\circ$ is an entry-wise multiplication operator. We denote by $p$ a pixel in a training image $\mathbf x$,  and $\mathbf{PL}_{\mathbf{x}}(p)$ its pseudo-label. As we see in (\ref{eq:contrastive_reconstruction_loss}), if the pixel $p$ belongs to a normal sample, the loss corresponding to this pixel is proportional to $\left(\hat{p} - p\right)^2$. To minimize $\left(\hat{p} - p\right)^2$, the model should map the output close to the input. If the pixel $p$ belongs to an anomalous  image and is labeled as anomalous, the corresponding loss is proportional to $ -\left(\hat{p} - p\right)^2$. To minimize it, the model should map the ouput away from the input.  With such a loss, the model reconstructs normal pixels well and reconstructs anomalous pixels badly. To minimize the loss, mini-batch gradient descent methods can be employed. 
	
	After the reconstruction model is updated, the new reconstruction residual images can be obtained, and the new pseudo-labels can be produced using these reconstruction residual images, and then the training samples can be refreshed to update the model again. The pseudo-label producing stage and model updating stage are iteratively executed in the manner of self-training so that the quality of the pseudo-labels and the performance of the model are gradually improved.
	
	\subsection{Overall Algorithm of KIST}
	We present the pseudocode of KIST in Algorithm \ref{alg:KIST}. It is worth noting that in practice the model $G$ would converge after a small number of iterations and the total number of iterations $I$ can take a small value such as $I=5$. 
	\begin{algorithm} 
		\caption{KIST} 
		\label{alg:KIST} 
		\renewcommand{\algorithmicensure}{\textbf{Output:}}
		\begin{algorithmic}[0]
			\STATE \textbf{Initialization} normal training samples $X_n$, anomalous training samples $X_{an}$, neural network $G$ with parameters $\Theta$, number of iterations $I$, fuzzy rule-represented knowledge;
			\STATE \textbf{1 Initial model constructing} 
			\STATE Optimize the parameters $\Theta$ of $G$ using normal samples $X_n$;
			\STATE \textbf{2 Model updating}
			\FOR{$i$ in $\{1,\cdots,I\}$}
				\STATE Obtain $T_e$ according to (\ref{eq:ts});
				%			\STATE Initialize $\mathbf{PL}_{\mathbf{X}_{an}}=\mathbf{0}$ having the same shape as $\mathbf{X}_{an}$;
				\STATE \textbf{2.1 Pseudo-label producing}
				\FOR{$\mathbf{x}$ in $X_{an}$}
					\STATE Get the anomalous-like region set $R_{\mathbf{x}}$ based on the knowledge according to (\ref{eq:find_an_regions}) and (\ref{eq:update_an_set});
					\STATE Get the pseudo-label $\mathbf{PL}_{\mathbf{x}}$ according to (\ref{eq:pseudo_label});
				\ENDFOR
				\STATE \textbf{2.2 Model updating using pseudo-label}
				\STATE Update the parameters $\Theta$ of $G$ using $X_n$ and $X_{an}$ along with the pseudo labels by minimizing the contrastive-reconstruction loss $L_{\text{cr}}$ (\ref{eq:contrastive_reconstruction_loss});
			\ENDFOR
			\STATE \textbf{Output} reconstruction model $G(\cdot;\Theta)$. \\
		\end{algorithmic} 
	\end{algorithm}
	\begin{remark}\label{rm:cr_loss}
		Generally, the knowledge given by experts can not cover all possible anomalies, so that some types of anomalies may not be mentioned in any of the rules. In the pseudo-label producing stage, we use the fuzzy rules to calculate the anomaly grade of a given region in an anomalous image. If a region is anomalous and the type of its anomaly is not involved in any rule, the region will have a low anomaly grade and the pixels of it will be mislabeled as normal.  On the other hand, if a region in an anomalous image is assigned a high anomaly grade by the fuzzy rules, the region is very likely to be anomalous and the pixels of this region will be labeled as anomalous. The two cases indicate that, if a pixel is labeled as anomalous, we can consider this pixel being anomalous with high confidence, however, we would better not trust a pixel in an anomalous sample labeled as normal is truly normal. Thus, in (\ref{eq:contrastive_reconstruction_loss}), for anomalous samples, only the pixels that are labeled as anomalous are used to update the reconstruction model.
	\end{remark}
	
	\begin{remark}\label{rm:set_alpha}
		When producing pseudo-labels, we need to determine the anomaly grade threshold $\alpha$. As we see in (\ref{eq:pc_irule})-(\ref{eq:ms_an}), for a given region $\mathbf{r}$ of an anomalous image $\mathbf{x}_i$, the anomaly grade $g^{(i)}(\mathbf{r})$ of $\mathbf{r}$ obtained according to the $i$th fuzzy rule is proportional to the truth value $tv^{(i)}$ of the rule. The truth value of  a rule can be determined based on the experience of the experts, however, different experts might give different truth values for one rule, which affects the values of $g(\mathbf{r})$'s and makes it hard to choose an appropriate anomaly grade threshold $\alpha$.  As we see, if the threshold $\alpha$ is low, some normal regions in the anomalous samples might be wrongly considered as anomalous. If $\alpha$ is  high, some anomalous regions may be misidentified as the normal ones. According to Remark \ref{rm:cr_loss},  we treat the pixels labeled as anomalies as truly anomalous, and for anomalous samples we only use pixels labeled as anomalous to update the reconstruction model. Since a relatively high anomaly grade threshold $\alpha$ can increase the possibility of pixels labeled as anomalies being truly anomalous, in practice a relatively high $\alpha$ is recommended when it is hard to determine $\alpha$.
	\end{remark}

	\subsection{Guided Filter-Based Post Processing}\label{sec:gfpost}
	
	In the testing phase, given an image $\mathbf{x}$, we could obtain the reconstruction image $\hat{\mathbf{x}}=G(\mathbf{x})$ and the residual image $e(\mathbf{x})=|\hat{\mathbf{x}} - \mathbf{x}|^2$. 
	Then, we expect to localize anomalous regions based on $e(\mathbf{x})$. In practice, the images could be contaminated with noise, which might lead to small fluctuations of residuals and thus affect the localization performance. We consider using guided filter~\cite{GuidedIF} to post process the residual image to mitigate the negative impact of the noise.
	
	When smoothing $e(\mathbf{x})$ to reduce the small fluctuations of residuals, we expect to obtain the filtering output $\mathbf{q}$ that preserves the edge information of the anomalies in the image $\mathbf{x}$. Therefore, we consider $\mathbf{q}$ as a linear transform of $\mathbf{x}$ in a window $\omega_{k}$ centered at the pixel $k$:
	\begin{equation}\label{eq:local_linear_model}
		q_i = a_k x_i + b_k, \forall i \in \omega_{k},
	\end{equation}
	where $(a_k, b_k)$ are linear coefficients of $\omega_k$. It is inferred from (\ref{eq:local_linear_model}) that $q_i$ has an edge only if $x_i$ has an edge, because $\Delta q_i=a\Delta x_i$. To determine the confidence $(a_k, b_k)$, we need the constraints from the filtering input $e(\mathbf{x})$. Specifically, we model $q_i$ as $e_i$ subtracting the unwanted components $n_i$ as 
	\begin{equation}
		q_i = e_i - n_i. \nonumber
	\end{equation}
	To smooth the small fluctuations while retaining the edge information in $e_i$, we seek $(a_k, b_k)$ that minimize the difference between $q_i$ and $e_i$ while maintaining linear model (\ref{eq:local_linear_model}). In the window $\omega_k$, we determine $(a_k, b_k)$ by minimizing the following cost function
	\begin{equation}\label{eq:min_linear_model}
		E(a_k, b_k) = \sum_{i\in \omega_k} \left((a_k x_i + b_k - e_i)^2 + \epsilon a_k^2 \right),
	\end{equation}
	where $\epsilon$ is a regularization parameter penalizing large $a_k$. 
	It is seen that (\ref{eq:min_linear_model}) is the linear ridge regression model~\cite{AppliedReg, TheElements} and its solution is given by
	\begin{gather}
		a_{k}=\frac{\frac{1}{|\omega|} \sum_{i \in \omega_{k}} x_{i} e_{i}-\mu_{k} \bar{e}_{k}}{\sigma_{k}^{2}+\epsilon} \nonumber \\
		b_{k}=\bar{e}_{k}-a_{k} \mu_{k} \nonumber
	\end{gather}
	where $\mu_k$ and $\sigma_k$ are the mean and variance of $x_k$, $\left|\omega\right| $ is the number of pixels in $\omega_k$ and $\bar{e}_k=\frac{1}{|\omega|} \sum_{i \in \omega_{k}} e_i$ is the mean of $e$ in $\omega_{k}$. After obtaining the linear coefficient $(a_k,b_k)$ of the window $\omega_k$, we can acquire the filtering output $q_i$ according to (\ref{eq:local_linear_model}). However, a pixel $i$ could be involved in different overlapping $w_k$ and $q_i$ in different $\omega_{k}$ may be not identical. A simple but intuitive strategy is averaging all the possible values of $q_i$. Specifically, after computing $(a_k, b_k)$ for all windows $\omega_k$, we could obtain the filtering ouput $q_i$ as follows:
	\begin{gather}
		q_i =\frac{1}{|\omega|}\sum_{k|i\in \omega_k}(a_k x_i+b_k) = \bar{a}_i x_i + \bar{b}_i,\nonumber
	\end{gather}
	where $\overline{a}_i=\sum_{k\in \omega_k}a_k$ and $\overline{b}_i=\sum_{k\in \omega_k}b_k$ are the average coefficients of all windows containing $i$. 

\section{Experiments}
	In this section, we first describe the datasets used in the experiment. 
	Then, we briefly describe the baseline and other advanced reconstruction-based anomaly localization methods for performance comparison.
	To fairly compare the KIST with those methods, we introduce two widely used evaluation metrics in the field of anomaly localization.
	The implementation details of the KIST method are given next.
	Finally, we give the qualitative and quantitative results to demonstrate the advantages of the KIST method over the counterparts.
	
%	Then, we introduce the implementation details of the KIST method. 
%	Besides, we briefly describe the baseline and other advanced reconstruction-based anomaly localization methods for performance comparison. To fairly compare these methods, we introduce two widely used evaluation metrics in the field of anomaly localization. Finally, we give the qualitative and quantitative results to demonstrate the advantages of the KIST method over the counterparts.
	
	\subsection{Experimental Setup}
	
	\subsubsection{Datasets description}
	
	In different industrial scenarios, anomalies can vary a lot. We conduct the experiment on three datasets to explore the generality of the KIST. The anomalies of these datasets vary in shape, size or color. A brief introduction of these datasets is as follows.
	
	Magnetic tile defect dataset (MTD), a real image dataset for surface detection, is provided by Huang~\cite{MTD}. The dataset contains 949 normal images and 289 anomalous images of region of interest of cropped magnetic tile. 
	KolektorSDD2 dataset (Kole) is constructed from images of defected production items which are provided by Kolektor Group d.o.o.~\cite{Kole}. The images are captured in a controlled industrial environment. The dataset is composed of 2979 normal images and 356 anomalous images with various types of defects (scratches, spots, etc.).
	MVTec dataset is published by MVTec Software Gmbh~\cite{MVTec}. The dataset is composed of 5 texture image categories and 10 object image categories. Totally 4096 normal images and 1258 anomalous images are available.

		\subsubsection{Methods for comparison}
	We compare the KIST with the baseline and other advanced reconstruction-based methods including SSIM \cite{SSIMAE},  WFD \cite{WFDL}, RIAD \cite{RIAD} and  MemAE \cite{MemAE}. The baseline use normal samples to construct the model by minimizing the loss given by (\ref{eq:l2_loss}). The SSIM, WFD and RIAD method utilize better loss functions to achieve performance gains. 
	The MemAE method constrains the features into a limited number of vectors to restrain the model's ability of reconstructing the anomalous samples and thus enhance the localization performance.
	
	\subsubsection{Evaluation metrics}
	
	To evaluate the anomaly localization performance, we adopt two widely used metrics, including area under the receiver operating characteristic curve (AUROC) and area under the per-region overlap curve (AUPRO). In the following  we briefly introduce these two metrics.
	
	AUROC is a single scalar value that measures the performance of a binary classifier. Given a residual threshold $t$, the true positive rate (TPR) is the proportion of correct predictions for anomalous pixels, and the false positive rate (FPR) is the proportion of incorrect predictions of anomalous pixels.
	By scanning over the range of residual thresholds, sorted serial values of TPR and FPR are obtained, and then AUROC is given as  
	\begin{equation*}
		AUROC=\sum_{i=1}TPR_i\cdot \Delta(FPR_i),
	\end{equation*}
	where $TPR_i$ and $FPR_i$ denote the $TPR$ and $FPR$ at the $i$-th sorted position, respectively, and $\Delta$ denotes the differential operation. AUPRO is another threshold-independent evaluation metric, which weights ground-truth regions of different size equally. Given a residual threshold $t$, we compute the percentage of correctly predicted pixels for each annotated anomalous regions $G_k$. The average over all $K$ anomalies yields
	\begin{equation*}
		PRO=\frac{1}{K}\sum_{k=1}^K\frac{\mathrm{Union}_k}{G_k} =\frac{1}{K}\sum_{i=1}^K\frac{P\cap G_k}{G_k},
	\end{equation*}
	where $P$ is the binary prediction given by $t$.
	We increase $t$ and compute PRO until FPR reaches 30\%. Then, we compute the area under PRO curve and normalize it to get AUPRO. 

	\subsubsection{Implementation details of the KIST} \label{Sec: ImplementKIST}
	\begin{table*}[t]
		\centering
		\caption{Properties and fuzzy rules used for knowledge representation.}
		\begin{tabular}{lllll}
			\toprule
			Properties of a region& \, & Fuzzy rules used on Kole and MVTec    & \, & Fuzzy rules used on MTD\\
			\midrule
			$\begin{aligned}
				&\text{Area~} \mathcal{V}_a(\mathbf{r}) = \frac{\Vert \mathbf{r} \Vert_0}{\Vert \mathbf{x} \Vert_0} \nonumber\\
				&\text{Gray~} \mathcal{V}_g(\mathbf{r}) = Mean(\mathbf{r}) \nonumber\\
				&\text{Shape~} \mathcal{V}_s(\mathbf{r}) = BCS(\mathbf{r}) \nonumber\\
				&\text{Unevenness~} \mathcal{V}_u(\mathbf{r}) =  Std(\mathbf{r}) \nonumber\\
				&\text{Symmetry~} \mathcal{V}_y(\mathbf{r}) = \frac{\Vert \mathbf{PL}_{\mathbf{b}_l} \wedge \mathbf{PL}_{\mathbf{b}_r'} \Vert_0}{\Vert \mathbf{PL}_{\mathbf{b}_l} \vee \mathbf{PL}_{\mathbf{b}_r'}\Vert_0} \nonumber
			\end{aligned}$  
			& \, &
			$\begin{aligned}
				&\text{IF $\mathcal{V}_a(\mathbf{r})$ is $F_h$ AND $\mathcal{V}_g(\mathbf{r})$ is $F_l$}\\
				&\text{THEN $\mathbf{r}$ is $F_{an}$}(tv=1)\\
				&\text{IF $\mathcal{V}_g(\mathbf{r})$ is $F_l$ AND $\mathcal{V}_s(\mathbf{r})$ is $F_h$}\\
				&\text{THEN $\mathbf{r}$ is $F_{an}$}(tv=0.8)\\
				&\text{IF $\mathcal{V}_a(\mathbf{r})$ is $F_l$ AND $\mathcal{V}_g(\mathbf{r})$ is $F_l$}\\
				&\text{THEN $\mathbf{r}$ is $F_{an}$}(tv=1)
			\end{aligned}$
			&  \, &
			$\begin{aligned}
				&\text{IF $\mathcal{V}_a(\mathbf{r})$ is $F_h$ AND $\mathcal{V}_g(\mathbf{r})$ is $F_l$ AND $\mathcal{V}_u(\mathbf{r})$ is $F_l$}\\
				&\text{THEN $\mathbf{r}$ is $F_{an}$}(tv=1)\\
				&\text{IF $\mathcal{V}_g(\mathbf{r})$ is $F_l$ AND $\mathcal{V}_s(\mathbf{r})$ is $F_h$ AND $\mathcal{V}_y(\mathbf{r})$ is $F_h$}\\
				&\text{THEN $\mathbf{r}$ is $F_{an}$}(tv=0.8)\\
				&\text{IF $\mathcal{V}_a(\mathbf{r})$ is $F_l$ AND $\mathcal{V}_g(\mathbf{r})$ is $F_l$ AND $\mathcal{V}_u(\mathbf{r})$ is $F_l$}\\
				&\text{THEN $\mathbf{r}$ is $F_{an}$}(tv=0.8)\\
				&\text{IF $\mathcal{V}_a(\mathbf{r})$ is $F_m$ AND $\mathcal{V}_g(\mathbf{r})$ is $F_h$ AND $\mathcal{V}_s$ is $F_h$}\\
				&\text{THEN $\mathbf{r}$ is $F_{an}$}(tv=1)
			\end{aligned} $
			\\
			\bottomrule
		\end{tabular}%
		\label{tab:ExpFuzzy}%
	\end{table*}%
	
	In the experiment, the ratio of the number of normal samples to the number of anomalous ones is set to 20:1, and some knowledge of the anomalies is available. Based on the training samples and the knowledge, the KIST method is implemented as follows.

	We first construct the initial reconstruction model using only normal samples. We adopt convolutional neural networks as the basic structures of the reconstruction model. All input images are scaled to the same size $256 \times 256$. 
	To obtain a more reliable initial reconstruction model, we use widely adopted image augmentation methods, such as random flip, random rotation, etc., to increase the diversity of normal  samples.
	To learn the parameters of the neural network, we use the mini-batch gradient descent method, and set the number of epochs to $200$, the batch size to $32$, and the learning rate to $1e-3$.

	After constructing the initial reconstruction model, we obtain residual images based on the model. Utilizing the residual images and the knowledge, we can produce pseudo-labels of anomalous samples for further training. To do so, we first use fuzzy rules to represent the knowledge. Given a region $\mathbf{r}$, we calculate its values of properties, namely, the area, gray, shape, uneven degree, and symmetry of $\mathbf{r}$, which are denoted by $\mathcal{V}_a(\mathbf{r}), \mathcal{V}_g(\mathbf{r}), \mathcal{V}_s(\mathbf{r}), \mathcal{V}_u(\mathbf{r}), \text{and }   \mathcal{V}_y(\mathbf{r})$, respectively. The computational formulas of all the properties are given in the first column of Table.~\ref{tab:ExpFuzzy}, where $Mean(\mathbf{r})$ and $Std(\mathbf{r})$ are the mean and standard deviation of the gray value of pixels of $\mathbf{r}$, $BCS(\mathbf{r})$ is the Boyce and Clark shape index~\cite{ShapeIndex} evaluating the shape of $\mathbf{r}$.
    We denote by $\mathbf{b}$ the minimum enclosing rectangle of $\mathbf{r}$, and denote by 
     $\mathbf{b}_l$ and $\mathbf{b}_r'$ left part and horizontal flip of right part of $\mathbf{b}$, respectively. The pseudo-labels of $\mathbf{b}_l$ and $\mathbf{b}_r'$ are denoted by $\mathbf{PL}_{\mathbf{b}_l}$ and $\mathbf{PL}_{\mathbf{b}_r'}$. 
     We use the IoU between $\mathbf{PL}_{\mathbf{b}_l}$ and $\mathbf{PL}_{\mathbf{b}_r'}$ to evaluate the symmetry $\mathcal{V}_y(\mathbf{r})$ of $\mathbf{r}$.
     Since the values of different properties have different scales, we standardize $\mathcal{V}_i(\mathbf{r})$ by 
	 \begin{equation}\label{eq:standard_property_value}
	 	\mathcal{V}'_i(\mathbf{r}) = \frac{\mathcal{V}_i(\mathbf{r})}{\gamma_i}, i \in\{a,g,s,u,m\}
	 \end{equation}
	 where $\gamma_i$ is the scale factor and can be determined based on the experience of the experts.
	 For example, the experts would know that the value $\mathcal{V}_i(\mathbf{r})$ of property $i$ greater than a number, say $v_h$, is high. If we scale $\mathcal{V}_i(\mathbf{r})$ approximately to $[0, 1]$, we would consider the standardized value $\mathcal{V}'_i(\mathbf{r})$ greater than $0.8$ to be high. Thus, we can take $0.8$ as the standardized value of $v_h$, i.e, $v_h  / \gamma_i=0.8$, and get $\gamma_i$. 
	 Using $\gamma_i$, we can standardize $\mathcal{V}_i(\mathbf{r})$ through (\ref{eq:standard_property_value}) and obtain each standardized value $\mathcal{V}'_i(\mathbf{r})$.
	 
\begin{figure}[t]
		\centering
		\hspace{-4mm}
		\includegraphics[width=0.75\columnwidth]{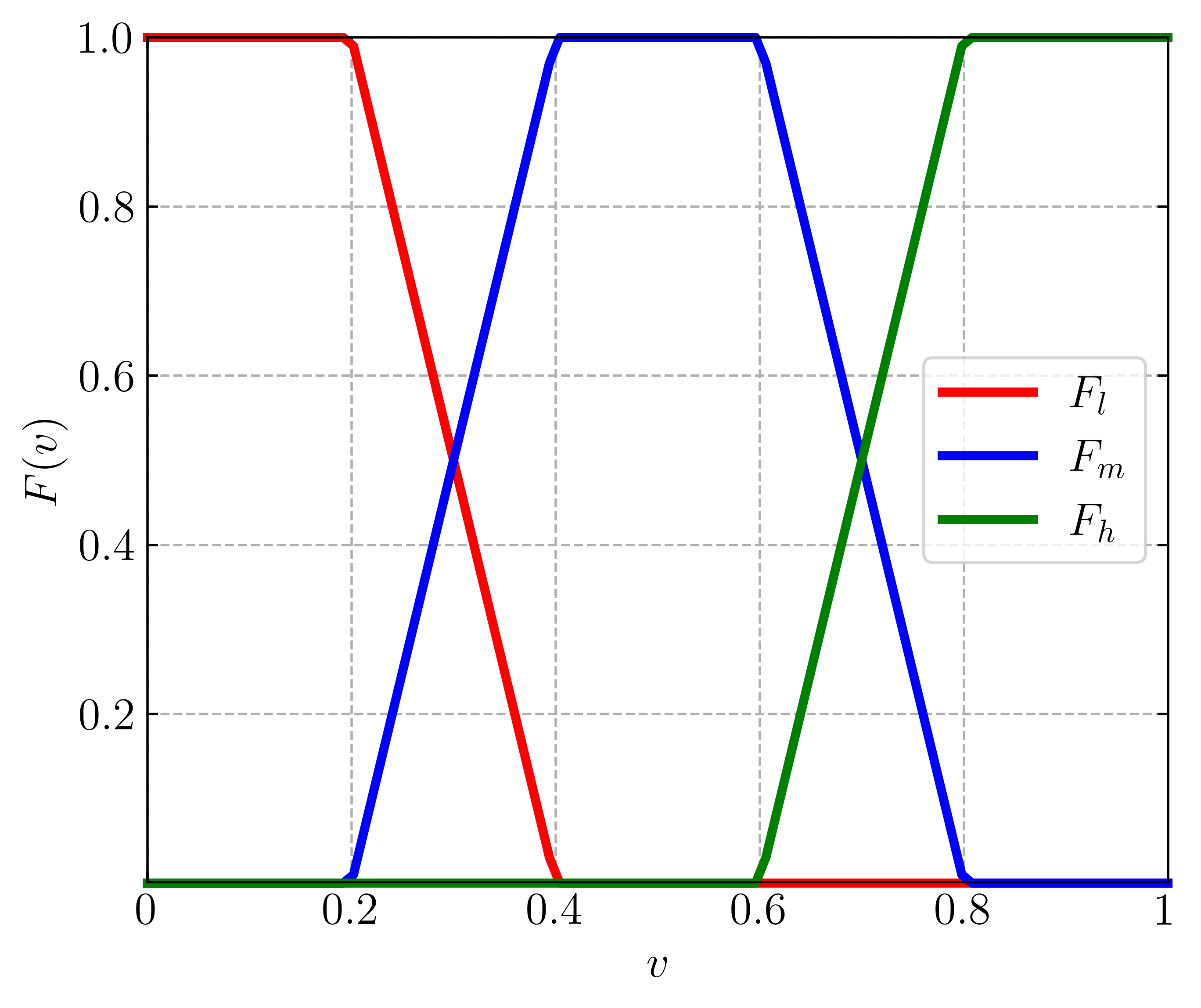}
		\caption{Trapezoidal membership functions (trapmfs) used for calculating the membership grades of the property value.}
		\label{Fig: trapmf}
	\end{figure}

	 After calculating the value of the properties $v_i = \mathcal{V}'_i(\mathbf{r}), i \in\{a,g,s,u,m\},$ of region $\mathbf{r}$, we evaluate the level of the value $v_i$ based on fuzzy predicates ``low",  ``mid" and ``high", which are denoted by fuzzy sets $F_l$, $F_m$ and $F_h$, respectively. With these fuzzy sets, the level of $v_i$ can be computed by membership functions $F_j(\cdot), j\in \{l, m, h\}$.
	 There are various kinds of membership functions available~\cite{mf}. Among them, the trapezoidal membership function (trapmf) is most popular~\cite{advantage_trapmf}. In the experiment, we use the trapmfs. The shape of the trampfs are shown in Fig.~\ref{Fig: trapmf}. We note that other shapes of the trapmf, or other membership functions can also be used. We evaluate the level of  each $v_i$ by calculating the membership grade $F_j(v_i)$.

	With the values $v_i$ of the properties and the fuzzy sets $F_j$, we can use fuzzy rules to represent the knowledge of anomalies. We note that some fuzzy rules can be applied in multiple scenarios, while certain scenarios require some specific fuzzy rules. For the Kole and MVTec (leather and wood) datasets, we use the fuzzy rules shown in the second column of Table~\ref{tab:ExpFuzzy}. For the MTD dataset, we adopt the fuzzy rules in the third column of Table~\ref{tab:ExpFuzzy}. 
	
	With the fuzzy rules, we use fuzzy rule-based pseudo-label producing module to generate pixel-wise pseudo-labels for anomalous samples. We traverse the residual threshold $t\in T_e$ to produce pseudo-labels. In $T_e$, we set the parameter $s$ to $0.3$, and thus the total number of steps is $10$. For the reason given in Remark \ref{rm:set_alpha}, we take a relatively high value $0.8$ for the anomaly grade threshold $\alpha$.
	Based on the generated pseudo-labels, we update the reconstruction model by minimizing contrastive-reconstruction loss (\ref{eq:contrastive_reconstruction_loss}). Since both reconstructing the normal pixels well and reconstructing the anomalous pixels badly are important, we assign the same weight to the two goals by setting $\lambda=1$ in  (\ref{eq:contrastive_reconstruction_loss}). 
	The pseudo-label producing stage and model updating stage are iteratively executed in the manner of self-training.
	
	In the testing phase, we test an image $\mathbf{x}$ based on its residual image $e(\mathbf{x})$. To smooth the small fluctuations of $e(\mathbf{x})$, we post process $e(\mathbf{x})$ using the guided filter. We use the original image $\mathbf{x}$ as the guided image, and set the radius $w$ of box window to $16$ and regularization parameter $\epsilon$ to $0.001$.

%	\subsection{Influence of Pretrained Models}
	\begin{figure}[t]
	\subfloat[]
	{
		\hspace{-5mm}
		\includegraphics[width=\columnwidth]{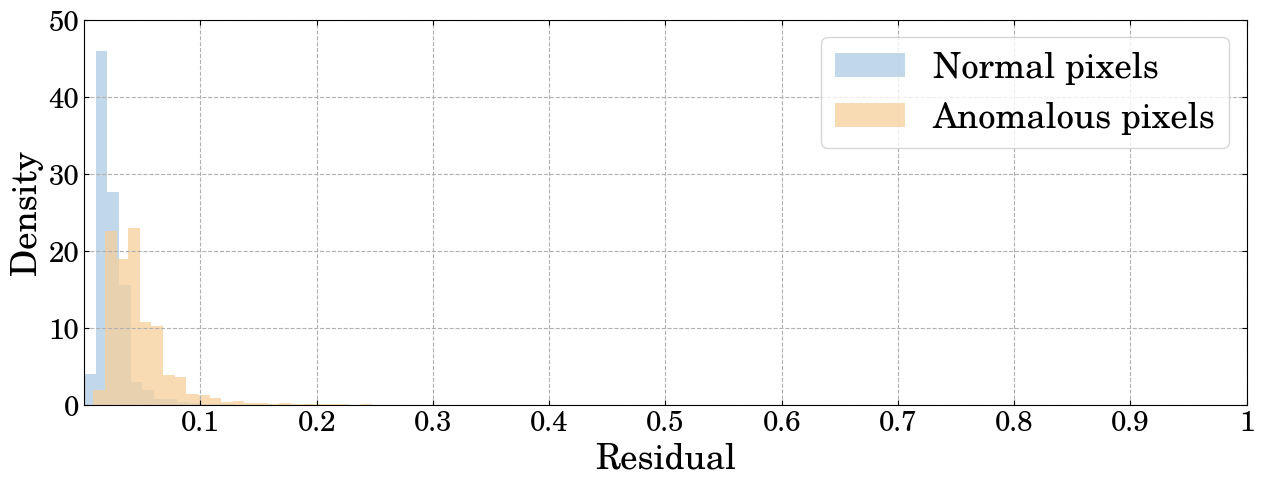}
		\label{fig:resid_dist_baseline}
	}
	\newline
	\subfloat[]
	{
		\hspace{-5mm}
		\includegraphics[width=\columnwidth]{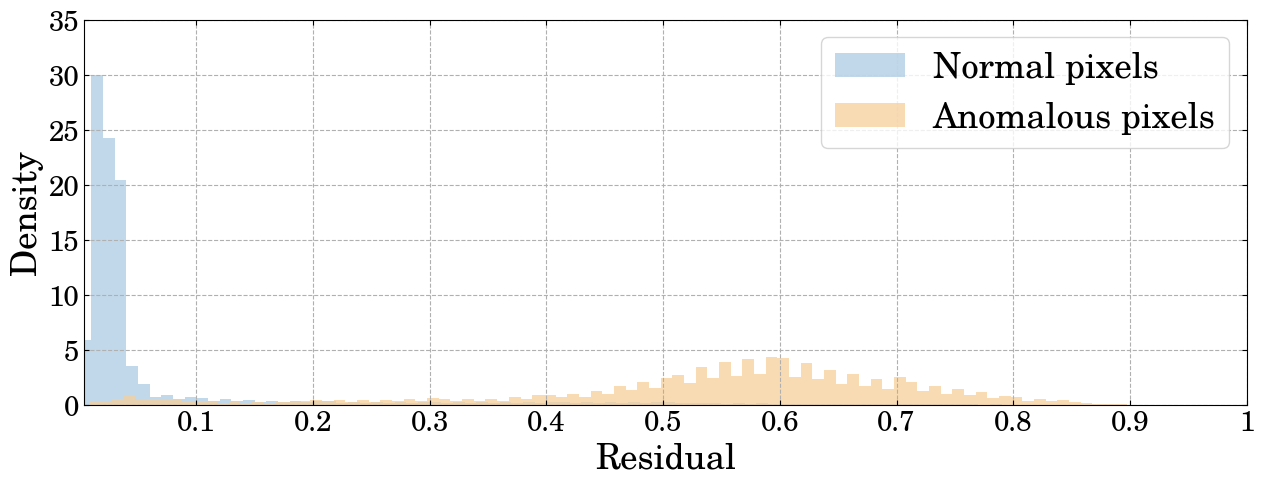}
		\label{fig:KIST}
	}
	\caption{Distributions of the residuals of the anomalous and normal pixels: (a) the case of the baseline, and (b) the case of the KIST method.}
	\label{Fig:residual_histogram}
	\end{figure}
	\subsection{Experimental Results}

	\subsubsection{Qualitative results}

		We localize the anomalies based on the residual images. If the residuals of the normal pixels and anomalous pixels are clearly separated, we can achieve good localization performance. We draw the distributions of the residuals obtained based on the KIST method and the baseline to show that the KIST can better separate the residuals of the anomalous pixels and the normal ones. The distributions of the residuals obtained based on the two methods are demonstrated in Fig.~\ref{Fig:residual_histogram}. As we see, the residuals of the normal and anomalous pixels generated by the baseline are not clearly separated, while the separation of the normal and anomalous pixels in the KIST method is more obvious.
		
%		We localize the anomalies based on the residual images. If the residuals of the normal  pixels and anomalous pixels are clearly separated, we can easily choose a residual threshold to distinguish the anomalous pixels from the normal ones. We draw the distributions of the residuals obtained based on the KIST method and the baseline to show that the KIST can better separate the residuals of the anomalous pixels and the normal ones. The distributions of the residuals obtained based on the two methods are demonstrated in Fig.~\ref{Fig:residual_histogram}. As we see,  it can be hard to find an appropriate residual threshold  to separate the residuals of the normal and anomalous pixels generated by the baseline, while the separation of the normal and anomalous pixels in the KIST method can be a lot easier.
		
		With the better separation of the normal and anomalous pixels, the KIST method can achieve better localization performance. A few localization results of the two methods are shown in Fig.~\ref{fig:test_result}. A residual threshold  $\mu_e + 2\sigma_e$ is used in both methods for the binarization of the residual images, where $\mu_e$ and $\sigma_e$ are the mean and standard deviation of the residuals of all the pixels of the normal training samples in the corresponding dataset. Since the residuals of the normal and anomalous pixels generated by the KIST are quite separated, 
%		the performance of the KIST is not sensitive to the choice of the residual threshold. 
		the residual threshold can be taken in a wide range for the KIST.
		On the other hand, the performance of the baseline is quite sensitive to the choice of the residual threshold. We carefully choose a suitable threshold  $\mu_e + 2\sigma_e$ for the baseline. It should be noted that it can be very hard to choose an appropriate threshold for the baseline in practice as the residuals of the normal and anomalous pixels generated by the baseline are not well separated. With the selected threshold $\mu_e + 2\sigma_e$, the baseline still  misidentifies some normal regions far from the truely anomalous regions as anomalies, and may fail to cover the entire anomalous regions. Compared with the baseline, the KIST can better localize the anomalies.
		
%		This residual threshold is chosen because normally the residuals of anomalous pixels are large. Trying to avoid missing anomalous pixels, we do not use a higher threshold like $\mu_e + 3\sigma_e$. With a relatively low residual threshold, some scattered normal pixels may be misidentified as anomalous, while such mistakes could be fixed in the post-processing procedure.
%		According to the distributions of the residuals obtained based on the two methods, adopting this residual threshold for the separation of the normal and anomalous pixels is appropriate.

%		\newcommand{\dataset}{MyMTD_95}
%		\newcommand{\name}{Class1}
		\newcommand{\crtline}{
			\subfloat 
			{
				\includegraphics[width=\wid in]{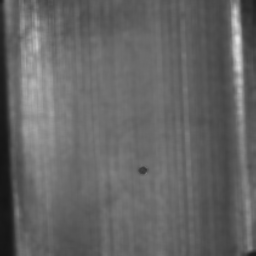}
				\hspace{\hsp mm}
			}
			\subfloat
			{
				\includegraphics[width=\wid in]{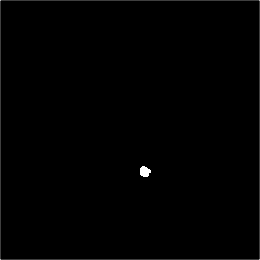}
				\hspace{\hsp mm}
			}
			\subfloat
			{
				\includegraphics[width=\wid in]{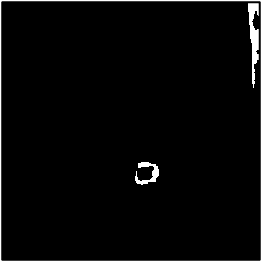}
				\hspace{\hsp mm}
			}
			\subfloat  
			{
				\includegraphics[width=\wid in]{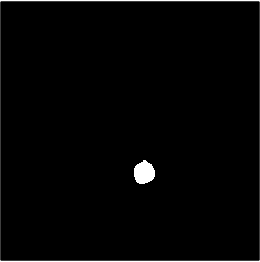}
			}
		}

		\begin{figure*}[htbp]    % 常规操作\begin{figure}开头说明插入图片
				% 后面跟着的[htbp]是图片在文档中放置的位置，也称为浮动体的位置，关于这个我们后面的文章会聊聊，现在不管，照写就是了
				\centering            
				\captionsetup[subfloat]{labelsep=none,format=plain,labelformat=empty}
				\newcommand{\img}{Fig71.png}
				\newcommand{\wid}{1.2}
				\newcommand{\hsp}{3}
				\newcommand{\vsp}{0}
				\crtline
				\newline
				\renewcommand{\img}{Fig72.png}
				\vspace{\vsp mm}
				\crtline
				\newline
				\renewcommand{\img}{Fig73.png}
				\vspace{\vsp mm}
				\crtline
				\newline
				\renewcommand{\img}{Fig74.png}
				\vspace{\vsp mm}
				\crtline
				\newline
				\renewcommand{\img}{Fig75.png}
				\vspace{\vsp mm}
				\crtline
				\newline
				\renewcommand{\img}{Fig76.png}
				\vspace{\vsp mm}
				\crtline
				\newline
				\renewcommand{\img}{Fig77.png}
				\vspace{\vsp mm}
				\subfloat[(a) original image]
				{
					\includegraphics[width=\wid in]{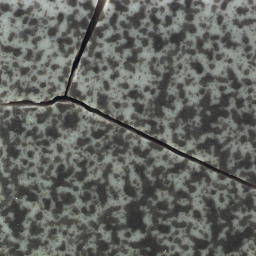}
					\hspace{\hsp mm}
				}
				\subfloat [(b) ground truth]
				{
					\includegraphics[width=\wid in]{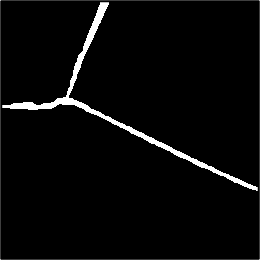}
					\hspace{\hsp mm}
				}
				\subfloat[(c) binary image (baseline)]
				{
					\includegraphics[width=\wid in]{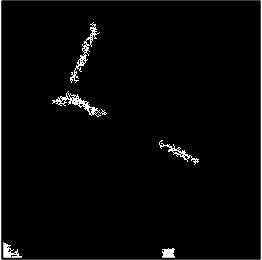}
					\hspace{\hsp mm}
				}
				\subfloat [(d) binary image (KIST)]
				{
					\includegraphics[width=\wid in]{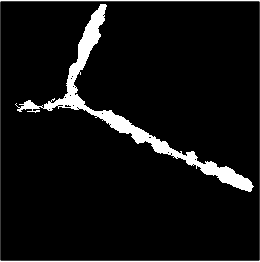}
				}
				\hspace{12mm}
				\caption{Visual comparisons of the KIST with the baseline method. Figures (a), (b), (c), and (d) are the testing images, the ground truth, the localization results of the baseline and KIST, respectively.}
				\label{fig:test_result}            
			\end{figure*}
		
	\subsubsection{Quantitative results}
		In addition to qualitative comparison, we quantitatively compare the KIST with the baseline and the other advanced reconstruction-based anomaly localization methods to further demonstrate the good performance of the KIST.
	
%		In addition to qualitative comparison, we also present quantitative results to show the advantages of the KIST. 
%		We compare the KIST with the baseline and other advanced reconstruction-based anomaly localization methods to show the good performance of the KIST. 
%		Besides, we also exhibit the benefit of guided filter-based post-processing method.
%		Finally, we evaluate the pseudo-labels of each iteration to show the rationality of the KIST.

	To show the effectiveness and generality of the KIST, we compare it with the other advanced reconstruction-based anomaly localization methods and use two neural networks~(NN), namely CAE \cite{SSIMAE} and Unet \cite{Unet}, to carry out the experiment. We present the comparison results in Table.~\ref{tab:other_loss_AUROC} and Table.~\ref{tab:other_loss_pro}. 
	The results without post-processing~(w/o PP) and with guided filter-based post-processing~(default) are both given. 
	As we see, compared with the baseline, the counterparts can enhance the localization performance in general. Specifically, the FFL method significantly improves the localization performance on the MTD dataset, but the improvement on the Wood dataset is not very satisfactory. The RIAD method performs well on the Leather and Wood datasets, however, it does not show obvious advantages on the MTD and Kole datasets. The memAE method exhibits its benefits on each dataset, but the overall improvement is limited. Compared with the baseline and the other counterparts, the KIST shows better localization performance and can be well applied to different neural networks and datasets. Besides, as we see in Table.~\ref{tab:other_loss_AUROC} and Table.~\ref{tab:other_loss_pro}, the guided filter-based post-processing can improve the localization performance on different datasets in general. 
	Nevertheless, without the post-processing method, the KIST already achieves good performance.

		\begin{table}[t]
		\centering
		\caption{AUROC performance given by different reconstruction-based methods (\%).}
		\begin{tabular}{cccccc}
			\toprule
			Method & NN & MTD & Kole & Leather & Wood \\
			\midrule
			baseline (w/o PP) & \multirow{6}[0]{*}{CAE} & 83.8 & 77.5 & 75.1 & 73.0  \\
			SSIM (w/o PP) &  & 81.3 & 79.0 & 78.0 & 73.0 \\
			FFL (w/o PP)&  & 88.9 & 79.3 & 75.1 & 74.5 \\
			RIAD (w/o PP)&  & 78.7 & 71.4 & 87.1 & 74.8 \\
			MemAE (w/o PP)&  & 85.4 & 80.5 & 78.2 & 75.8 \\
			KIST (w/o PP)&  & \textbf{92.4} & \textbf{85.5} & \textbf{90.1} & \textbf{76.5} \\
			\midrule[0.5pt]
			baseline& \multirow{6}[0]{*}{CAE} & 85.2 & 81.3 & 79.8 & 74.4  \\
			SSIM&  & 84.3 & 82.0 & 83.1 & 75.4 \\
			FFL& & 90.1  & 83.4  & 78.2  & 77.2 \\
			RIAD&  & 79.6  & 76.5  & 88.7  & 75.8 \\
			MemAE & & 88.4  & 83.2  & 82.3  & 76.0 \\
			KIST& & \textbf{94.5} & \textbf{88.3} & \textbf{93.4} & \textbf{78.6} \\
			\midrule[0.5pt]
			baseline (w/o PP)& \multirow{5}[0]{*}{Unet} & 86.3 & 72.9 & 78.5 & 76.8 \\
			SSIM (w/o PP) & & 90.4 & 74.2 & 76.3 & 79.0 \\
			FFL (w/o PP) & & 93.7 & 81.6 & 88.5 & 80.3 \\
			RIAD (w/o PP) & & 80.1 & 70.1 &96.7 & 85.8 \\
			KIST (w/o PP)& & \textbf{94.5} & \textbf{93.2} & \textbf{97.8} & \textbf{86.8} \\
			\midrule[0.5pt]
			baseline & \multirow{5}[0]{*}{Unet} & 88.3  & 75.6  & 82.1  & 80.5 \\
			SSIM & & 91.1  & 76.3  & 79.3  & 82.1 \\
			FFL & & 93.9  & 84.5  & 89.2  & 83.3 \\
			RIAD & & 83.2  & 75.6  & 98.6  & 86.8 \\
			KIST & & \textbf{96.1} & \textbf{95.2} & \textbf{99.2} & \textbf{88.7} \\
			\bottomrule
		\end{tabular}
		\label{tab:other_loss_AUROC}
	\end{table}

		\begin{table}[t]
		\centering
		\caption{AUPRO performance given by different reconstruction-based methods (\%).}
		\begin{tabular}{cccccc}
			\toprule
			 Method & NN & MTD & Kole & Leather & Wood \\
			\midrule
			baseline & \multirow{6}[0]{*}{CAE} & 76.5 & 73.2 &  50.8 & 50.5 \\
			SSIM (w/o PP) &   & 76.4 & 71.5 & 56.1 & 60.5 \\
			FFL (w/o PP)&  & 82.4 & 72.0 & 75.3 & 61.3 \\
			RIAD (w/o PP) &  & 72.4 & 65.5 & 77.2 & 62.5 \\
			MemAE (w/o PP)&  & 77.8 & 74.7 & \textbf{77.9} & 53.5 \\
			KIST (w/o PP)&  & \textbf{83.2} & \textbf{80.5} & 72.3 & \textbf{65.0} \\
			\midrule[0.5pt]
			baseline& \multirow{6}[0]{*}{CAE} & 78.1  & 75.2  & 51.2  & 52.3 \\
			SSIM &  & 77.9  & 74.0    & 57.5  & 61.6 \\
			FFL &  & 84.2  & 73.2  & 76.3  & 65.4 \\
			RIAD &  & 76.5  & 68.5  & 78.2  & 64.2 \\
			MemAE & & 81.3  & 79.8  & \textbf{78.9}  & 56.7 \\
			KIST & & \textbf{84.6} & \textbf{82.5} & 74.4 & \textbf{66.5} \\
			\midrule[0.5pt]
			baseline (w/o PP) & \multirow{5}[0]{*}{Unet} & 70.1 & 70.5 & 65.4  & 60.8  \\
			SSIM (w/o PP) & & 74.2 & 71.5 & 65.2 & 61.2 \\
			FFL (w/o PP) & & 83.2 & 73.4 & 68.8 & 65.9 \\
			RIAD (w/o PP) & & 72.2 & 75.2 & 96.5 & 72.6 \\
			KIST (w/o PP) & & \textbf{83.4}  & \textbf{92.1} & \textbf{98.2} & \textbf{75.4} \\
			\midrule[0.5pt]
			 baseline &  \multirow{5}[0]{*}{Unet} & 74.2  & 73.4  & 66.4  & 61.2  \\
			 SSIM & & 80.3  & 73.5  & 66.8  & 60.9 \\
			FFL & & 85.1  & 74.8  & 72.2  & 66.8 \\
			RIAD & & 75.3  & 77.7  & 96.6  & 75.5 \\
			KIST & & \textbf{86.7}  & \textbf{92.5} & \textbf{98.5}  & \textbf{78.0} \\
			\bottomrule
		\end{tabular}
		\label{tab:other_loss_pro}
	\end{table}

\begin{figure}[t]    % 常规操作\begin{figure}开头说明插入图片
			% 后面跟着的[htbp]是图片在文档中放置的位置，也称为浮动体的位置，关于这个我们后面的文章会聊聊，现在不管，照写就是了
			\centering            % 前面说过，图片放置在中间
			%				\captionsetup[subfloat]{labelsep=none,format=plain,labelformat=empty}
			\subfloat[]{
				\hspace{-7mm}
				\label{fig:ep_ROC}\includegraphics[width=\columnwidth]{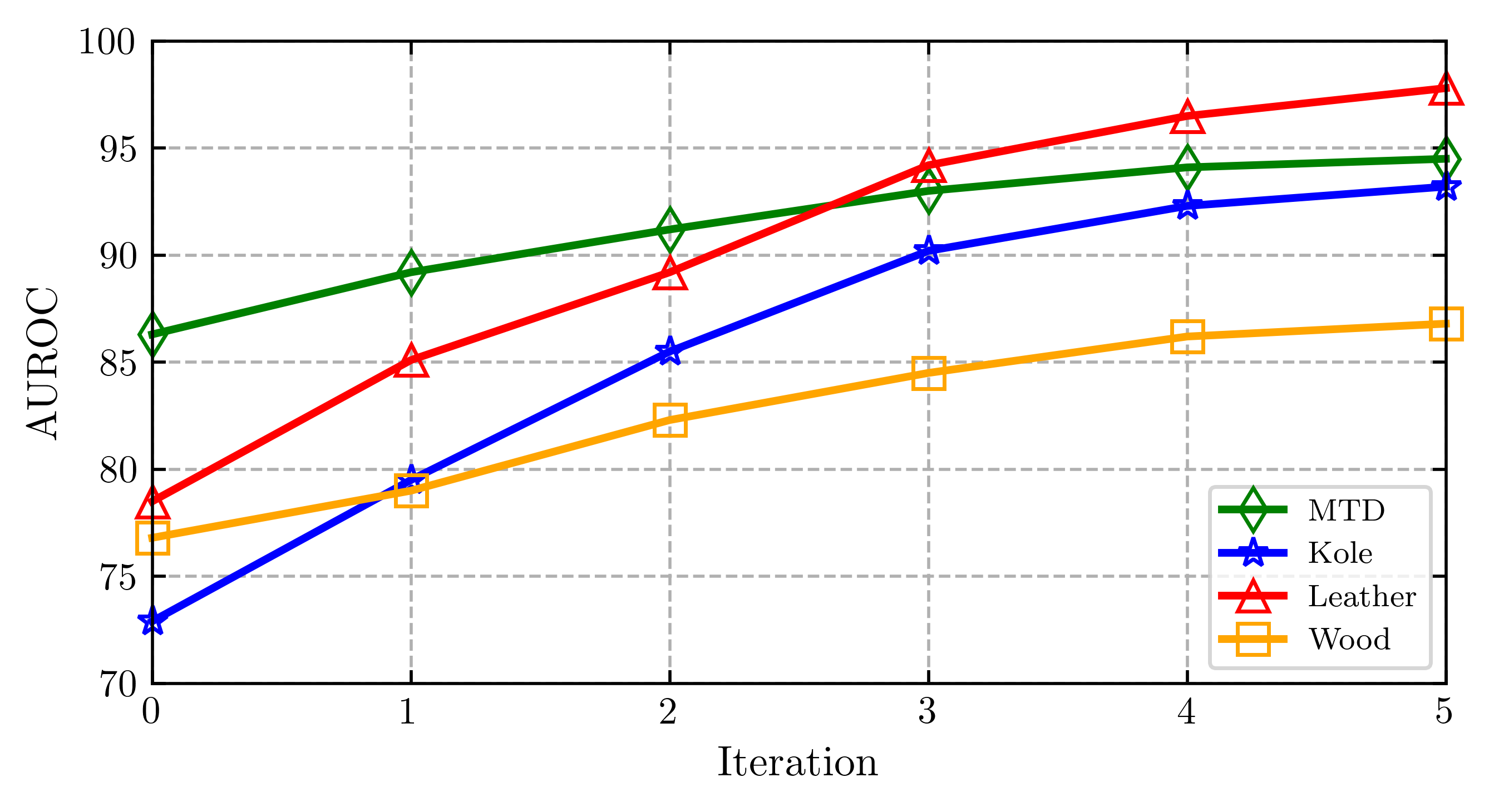}
			}
			\newline
			\vspace{-1mm}
			\subfloat[]{
				\hspace{-7mm}
				\label{fig:ep_PRO}\includegraphics[width=\columnwidth]{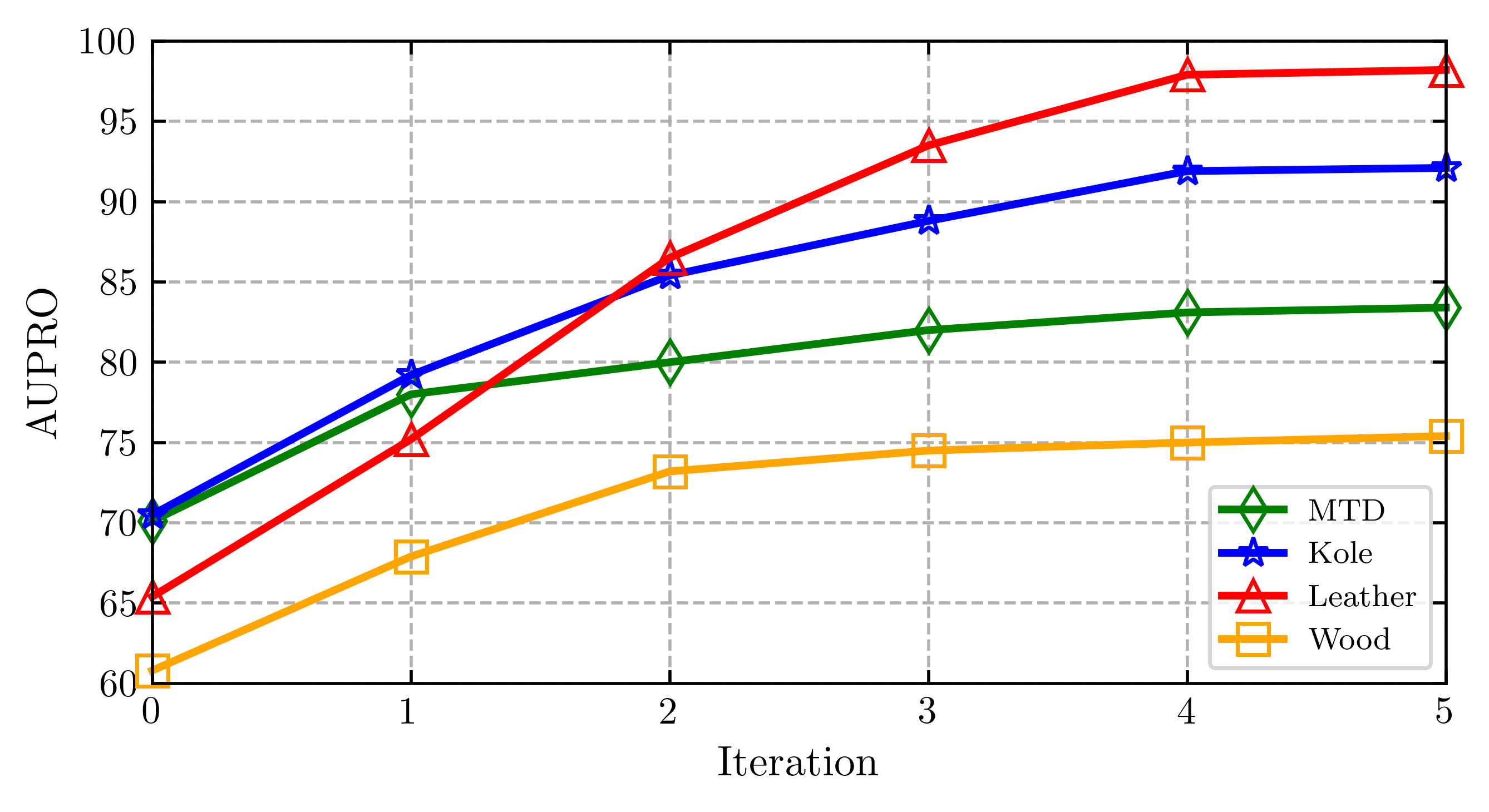}
			}
			\caption{Performance of the reconstruction model using Unet over iterations: (a) under the evaluation metric AUROC, and (b) under the metric AUPRO.}    % 整个图片的说明，注释写在{}内
			\label{fig:model_iterations}            % 整个图片的标签编号，注意这里跟子图是一样的道理，标签不能重复 
		\end{figure}

	We note that the KIST benefits from the manner of self-training.  In the KIST, the two stages of pseudo-label producing and model updating are iteratively conducted such that the quality of pseudo-labels and the performance of the reconstruction model are gradually improved. In Fig.~\ref{fig:model_iterations}, we demonstrate the performance of the model in each iteration. As we see, the performance of the model is gradually improved and becomes stable after a small number of iterations. It shows the effectiveness of the self-training approach.

\section{Conclusion}
In this paper, we propose a novel reconstruction-based anomaly localization method named KIST which integrates knowledge into reconstruction model through self-training. 
The KIST utilizes weakly labeled anomalous samples in addition to the normal ones and utilizes knowledge to produce pixel-level pseudo-labels of the anomalous samples. Based on the pseudo labels, the contrastive-reconstruction loss which promotes the reconstruction of normal pixels while suppressing the reconstruction of anomalous pixels is employed to update the reconstruction model.
In the experiment, we compare the KIST with the other reconstruction-based anomaly localization methods on different datasets. The experimental results confirm the advantages of the KIST. In the future, we are to explore more efficient ways of utilizing the knowledge to help improve the localization performance.

\ifCLASSOPTIONcaptionsoff
  \newpage
\fi

\end{document}